\documentclass[preprint,12pt]{elsarticle}

\usepackage{amssymb,amsmath,amsthm,amsfonts,mathtools,caption,subcaption,csquotes,enumitem,hyperref,makecell,booktabs,tabularx}
\usepackage{color}

\newcommand{\bx}{\mathbf{x}}
\newcommand{\by}{\mathbf{y}}

\newcommand{\bt}{\mathbf{t}}

\newcommand{\bv}{\mathbf{v}}

\newcommand{\bI}{\mathbf{I}}

\newcommand{\bK}{\mathbf{K}}

\newcommand{\bxi}{\boldsymbol{\xi}}

\newcommand{\btheta}{\boldsymbol{\theta}}

\newcommand{\calD}{\mathcal{D}}

\newcommand{\calZ}{\mathcal{Z}}

\newcommand{\calX}{\mathcal{X}}
\newcommand{\calY}{\mathcal{Y}}
\newcommand{\calU}{\mathcal{U}}

\newcommand{\E}[2]{\mathbb{E}_{#1}\left[#2\right]}           

\newcommand{\R}{\mathbb{R}}

\newcommand{\nothree}[1]{\left[\operatorname{NO}_3^-\right]}

\newcommand{\littletaller}{\mathchoice{\vphantom{\big|}}{}{}{}}

\theoremstyle{definition}

\newcommand{\ie}{\textit{i.e.}}
\newcommand{\eg}{\textit{e.g.}}
\newcommand{\etc}{\textit{etc}}
\newcommand{\wrt}{\textit{w.r.t.\,}}
\newcommand\restr[2]{{
  \left.\kern-\nulldelimiterspace 
  #1 
  \littletaller 
  \right|_{#2} 
  }}

\journal{Journal of Computational Physics}

\begin{document}

\begin{frontmatter}



\title{Bayesian Inverse Problems with Conditional Sinkhorn Generative Adversarial Networks in Least Volume Latent Spaces}


\author[inst1]{Qiuyi Chen}

\affiliation[inst1]{organization={Department of Mechanical Engineering, University of Maryland},
            city={College Park},
            postcode={20742}, 
            state={MD},
            country={USA}}

\author[inst2]{Panagiotis Tsilifis}
\author[inst1]{Mark Fuge}

\affiliation[inst2]{organization={Robotics, AI \& Computer Vision Group, General Electric Vernova Advanced Research},
            city={Niskayuna},
            postcode={12309}, 
            state={NY},
            country={USA}}

\begin{abstract}
Solving inverse problems in scientific and engineering fields has long been intriguing and holds great potential for many applications, yet most techniques still struggle to address issues such as high dimensionality, nonlinearity and model uncertainty inherent in these problems.
Recently, generative models such as Generative Adversarial Networks (GANs) have shown great potential in approximating complex high dimensional conditional distributions and have paved the way for characterizing posterior densities in Bayesian inverse problems, yet the problems' high dimensionality and high nonlinearity often impedes the model's training. 
In this paper we show how to tackle these issues with Least Volume\textemdash a novel unsupervised nonlinear dimension reduction method\textemdash that can learn to represent the given datasets with the minimum number of latent variables while estimating their intrinsic dimensions. Once the low dimensional latent spaces are identified, efficient and accurate training of conditional generative models becomes feasible, resulting in a latent conditional GAN framework for posterior inference. 
We demonstrate the power of the proposed methodology on a variety of applications including inversion of parameters in systems of ODEs and high dimensional hydraulic conductivities in subsurface flow problems, and reveal the impact of the observables' and unobservables' intrinsic dimensions on inverse problems. 
\end{abstract}

-
\begin{highlights}
\item The Least Volume autoencoder (LVAE) is introduced that is trained to explore low dimensional structures of a given dataset for unsupervised dimensionality reduction and manifold learning.
\item The key characteristics of the LVAE is that it minimizes a regularized loss function taking into account the volume of the dataset in the latent space and that a K-Lispchitz condition on the decoder guarantees convergence to a non degenerate solution.
\item A conditional Sinkhorn Generative Adversarial Network (GAN) is proposed for exploring Bayesian posterior densities.
\item Combining the LVAE and conditional Sinkhorn GAN results in an efficient framework for solving high dimensional inverse problems by encoding the data and training the GAN on the low dimensional latent spaces, and then decoding the solution back to the original space.
\item The proposed methodology is applied on two high dimensional problems, that of inferring the source of a nonlinear ODE system given partial output signal and the problem of subsurface permeability inference from oil production data in a two dimensional oil reservoir where both input and output quantities comprise a few thousands of dimensions.
\item The important correlation between the uncertainty in the posterior's prediction and the intrinsic dimensionality of the posterior's input condition is uncovered.
\end{highlights}


\end{frontmatter}


\section{Introduction}
\label{sec:intro}

Engineering design and optimization tasks almost always entail procedures of inference of quantities of interest or design parameters from indirect measurements of observables that are typically the outcome of a physical experiment or the output of a computational model~\cite{tarantola}. In the majority of cases, the observations available to the engineer are limited and the dependence of the design or input parameters on the observed output is described by a highly nonlinear physical process, making the inference task and the quantification of uncertainty around the input a nontrivial process. From a theoretical standpoint, numerous works in the literature have been dedicated to characterizing the well-posedness of the problem and determining the conditions under which a solution is feasible and can be numerically approximated. Such conditions vary from defining appropriate regularizers on a misfit function~\cite{engl, calvetti}, to ensuring Lipschitz conditions on the log-likelihood function in a Bayesian context~\cite{stuart, cotter_stuart}. From an applications perspective, major challenges are typically the computationally expensive forward model involved in the experiment, resulting in only a limited number of available observations and the high dimensionality of both input parameters and observables that makes it difficult for the inference algorithm to converge to a solution. Applications where such issues manifest include ocean dynamics~\cite{Herbei2008GyresAJ}, seismic inversion~\cite{russell}, numerical weather forecasting~\cite{ollinaho, ekblom}, ice-sheet flow~\cite{isaac, petra, petra_siam} and large chaotic dynamical systems~\cite{springer}.

When following a Bayesian paradigm, traditional techniques for inverse problems are highly associated with commonly performing Bayesian inference, that is the task of exploring a conditional distribution associated with our \emph{posterior} knowledge about the unknown parameters, after observations are collected. Several variations of Markov Chain Monte Carlo (MCMC) techniques~\cite{haario, marzouk_najm, bilionis_inv, tsilifis_rspa} have provided efficient ways of exploring complex posterior densities in high dimensional settings or on manifolds. Alternatives methods to MCMC include Variational Inference~\cite{hoffman, broderick, tsilifis_vi}, where posteriors approximated by choosing an optimal density from a family of distributions, therefore transforming the inference task from a sampling strategy to an optimization algorithm. Due to the fact that both approaches require a tremendous amount of computational effort, constructing surrogates has been one of the most common ways to accelerate the learning procedure of the posterior density. Popular choices for surrogate modeling are Polynomial Chaos~\cite{marzouk_jcp, marzouk_xiu, tsilifis_siamuq, tsilifis_vrvm, tsilifis_cs, tsilifis_wiener}, Gaussian Processes~\cite{kennedy_ohagan, bilionis_mo, tsilifis_cmame} or adaptive sparse grid collocation~\cite{ma_zabaras}. 

The continuously rising momentum of deep learning in engineering sciences has resulted in significant contributions in the solution of inverse problems in medical imaging and compressed sensing~\cite{jin, fan_jcp, mardani}, geosciences and reservoir engineering~\cite{laloy, mo_zabaras, mo_zabaras2}, aerodynamic design~\cite{ghosh_amse, chen_asme, tsilifis_scitech22} and other areas. The use of ``deep" architectures, referring to the number of layers in an (artificial) neural network involved in modeling the physical process or the inference procedure, allows for theoretically capturing arbitrarily high nonlinearities and enabling accurate learning of the functional representations, provided that sufficient amounts of data are available during training. In a forward uncertainty propagation setting, deep learning architectures have demonstrated great potential in modeling solution to elliptic PDEs~\cite{tripathy_bilionis}, flow through porous media~\cite{zhu_zabaras} and turbulence~\cite{geneva_zabaras}. Multi-fidelity settings have also been explored using physics informed deep learning~\cite{meng_babaee} or transfer learning~\cite{kontolati_scitech}, while inverse problems have been tackled by applying Bayesian \emph{physics-informed} neural networks {B-PINNS} for modeling posterior densities~\cite{yang_meng}. 

Deep Generative or \emph{pushforward} models is one of the most prominent fields of machine learning techniques for modeling conditional densities and addressing inverse modeling. Inspired by the seminal works in adversarial learning~\cite{goodfellow2014generative}, variational auto-encoders (VAE)~\cite{kingma2013auto} and normalizing flows~\cite{dinh2016density, dinh2014nice}, generative models have been modified to their conditional counterparts in order to account for condition variables and thus generate samples from conditional distributions learned from observed data. The key idea behind conditioning in generative models is to incorporate the observable quantity as a conditioning variable in the neural network architecture that describes generative-discriminative process in generative adversarial nets (GANs), the encoder-decoder framework in a VAE setting or in the coupling layers in a normalizing flow. Conditional VAEs have already found applications in molecular design~\cite{lee_cvae}, image reconstruction~\cite{zhang_cvae}, while an novel intrusive approach has also been developed for PDE-constrained inverse problems~\cite{damoulas}. Perhaps one of the main drawbacks with cVAEs is that the latent space and its dimensionality needs to be specified a priori by the experimenter and choosing the dimension to be low enough in order to achieve computational efficiency, while maintaining descriptive accuracy can be a hit-or-miss situation. Normalizing flows on the other hand, do not aim at achieving dimensionality reduction. The latent (normalized) variable is required to preserve the dimensionality of the original input variable in order to ensure invertibility in the mapping as guaranteed by the Independent Component Analysis theorem~\cite{hyvarinen}, which makes it inefficient in very high dimensional settings. Interesting applicability of normalizing flows and conditional variations of it can be found in radiology~\cite{dasgupta}, astrophysics~\cite{haldemann}, subsurface flow~\cite{govinda} and industrial gas turbine airfoil design~\cite{tsilifis_scitech22}. At last, GANs have played their own pivotal role in advancing computational efficiency in inverse modeling. In~\cite{patel_cmame}, the authors use a GANs for modeling the prior knowledge in a Bayesian and using it as a dimension reduction technique, performing the inference in the latent space. Other variations of conditional GANs have shown interesting potential in speech enhancement~\cite{qin} using the Wasserstein distance is used for solving the minmax problem. Additionally, a Lipschitz condition on the critic has been shown to further improve performance in parameter inference of PDEs~\cite{ray_cmame}, resulting a loss function regularized by a gradient penalty.

In this paper, we introduce the \emph{latent conditional Sinkhorn GANs (latent CSGANs)}, which break down the approximation of the target posterior distribution into two consecutive steps that not only increase efficiency but also provide useful information about the posterior. Specifically, the proposed framework first learns the low dimensional structure of the target posterior's support by encoding the target posterior's input and output to latent spaces, via invertible mappings, and \emph{automatically} reducing the latent spaces' dimensions as much as possible. Then, it minimizes the Sinkhorn divergence in order to approximate the posterior with a pushforward generative model in these low dimensional latent spaces, thus, accelerating and stabilizing training by alleviating the curse of dimensionality and by replacing the problematic minimax game from vanilla GANs. We use two high dimensional examples to show that this two-stage posterior inference paradigm can approximate complex posteriors accurately and efficiently. We also reveal the important correlation between the input condition's intrinsic dimension\textemdash estimated by the latent space's dimension\textemdash and the uncertainty around the posterior solution, which can allow us to assess the posterior's complexity even before retrieving it. 

The paper is structured as follows: The proposed methodology is presented in detail in \S\ref{sec:method}. More specifically, in \S\ref{sec:volume} we introduce the novel \emph{least volume autoencoders} that are used for dimensionality reduction as a special case of generic autoencoders (\S\ref{sec:auto}) with a regularization term in their loss function (\S\ref{sec:regulatization}) that accounts for minimizing the volume of the data when projected in the latent space. The resulting formulation and optimization problem are stated in \S\ref{sec:loss_function}. Next, the conditional Sinkhorn GANs are introduced in \S\ref{sec:csgan}. Section~\ref{sec:pushforward} introduced the generic concept of pushforward models while \S\ref{sec:sinkhorn_gans} introduced the type of GANs trained with the Sinkhorn divergence incorporated into their loss function. Then, \S\ref{sec:csgan_sub} introduced the conditional counterpart of the Sinkhorn GANs and at last, \S\ref{sec:lvae_csgan} describes how the latter is combined with the least volume autoencoder in order to explore the inverse problem solution on the reduced dimension latent space. The proposed methodology is illustrated with two numerical examples presented in \S\ref{sec:experiment}
where first (\S\ref{sec:ko_ode}) we apply the approach on a simple inversion problem consisting of inferring the initial conditions of an ODE system from observing a partial signal of its solution and next (\S\ref{sec:oilextraction}) we challenge our method on the high-dimensional inversion problem of log-permeability inference on a two-phase flow oil reservoir based on data that characterized the oil saturation of the domain the oil production curves are specific production wells located throughout the domain boundaries. Finally, we summarize our findings and propose future research directions in the conclusions section \S\ref{sec:conclusion}.

\section{Methodology}
\label{sec:method}
The approximation of high dimensional conditional distribution can be decomposed into two consecutive tasks\textemdash the retrieval of the low dimensional manifold structure that supports the data distribution, and the learning of the conditional distribution restricted to this low dimensional manifold. Correspondingly, we first introduce the \emph{Least Volume Autoencoder} that can \emph{automatically} capture the low dimensionality and nonlinearity of the data manifold and \emph{invertibly} transform the data into a low dimensional latent space for efficient analysis. Thereafter, we present the \emph{conditional Sinkhorn generative adversarial network} (CSGAN)\textemdash a generative model based on optimal transport\textemdash for approximating and sampling from complicated conditional distributions. At last, we show how to integrate them to learn conditional distributions in high dimensional spaces efficiently.

\subsection{Least Volume Autoencoders}
\label{sec:volume}
In this starting section, we introduce the manifold learning method that can retrieve the dataset's low dimensional nonlinear structure and concisely represent it to facilitate downstream applications.

\subsubsection{Autoencoders}
\label{sec:auto}
Assume $\calX$ is a set of points in the \emph{data space} $X$ that we denote as the \emph{dataset} $\calX \subset X$, following some probability distribution with density $p_{\calX}(x)$. 
An \emph{autoencoder} is an ordered pair $(e,g)$ consisting of two functions, namely an \emph{encoder} $e:X\to Z$ and a \emph{decoder} $g:Z\to X$, that are typically modeled using neural networks and are trained by minimizing a reconstruction loss function  
\begin{equation}
J(\btheta) := \mathbb{E}_{x\sim p_{\calX}(x)} [d(g_{\btheta} \circ e_{\btheta}(x), x)]   
\label{eq:rec}
\end{equation}
with respect to the parameters $\btheta$ on $g$ and $e$, where $Z$ is the \emph{latent space} and $d: X\times X\to \mathbb{R}$ is a user-specified distance function. Despite its simple formulation, allowing $g$ and $e$ to be \emph{continuous} functions with arbitrary complexity can reduce the reconstruction error to 0, thus implying that $g = e^{-1}$ and $e$ becomes a \emph{topological embedding} of the dataset $\mathcal{X}\subseteq X$, such that the \emph{latent set} $\mathcal{Z} \coloneqq e(\mathcal{X})\subseteq Z$ is homeomorphic to $\mathcal{X}$~\cite{qiuyi2024compressing}. Consequently, given the trained autoencoder, if we want to learn a continuous function $f:\mathcal{X}\to Y$ (or $f:Y \to \mathcal{X}$), we can equivalently learn its latent counterpart $f': \mathcal{Z}\to Y$ (or $f':Y \to \mathcal{Z}$) and recover $f$ via $f = f'\circ e \Leftrightarrow f' = f \circ g$ (or $f = g\circ f' \Leftrightarrow f' = e \circ f$). This equivalence is particularly beneficial when the learning cost of $f$ does not scale well with its dimensionality but $\dim Z \ll \dim X$, as learning $f'$ in $Z$ significantly mitigates the curse of dimensionality. 

Therefore, retrieving a $Z$ of dimension as small as possible is important for high dimensional problems. In what follows, we show how to obtain a continuous autoencoder $(e,g)$ that learns to reduce $\dim Z$ as much as possible. 

\subsubsection{Least Volume Regularization}
\label{sec:regulatization}

\begin{figure}[bt!]
    \centering
    \includegraphics[width=\textwidth]{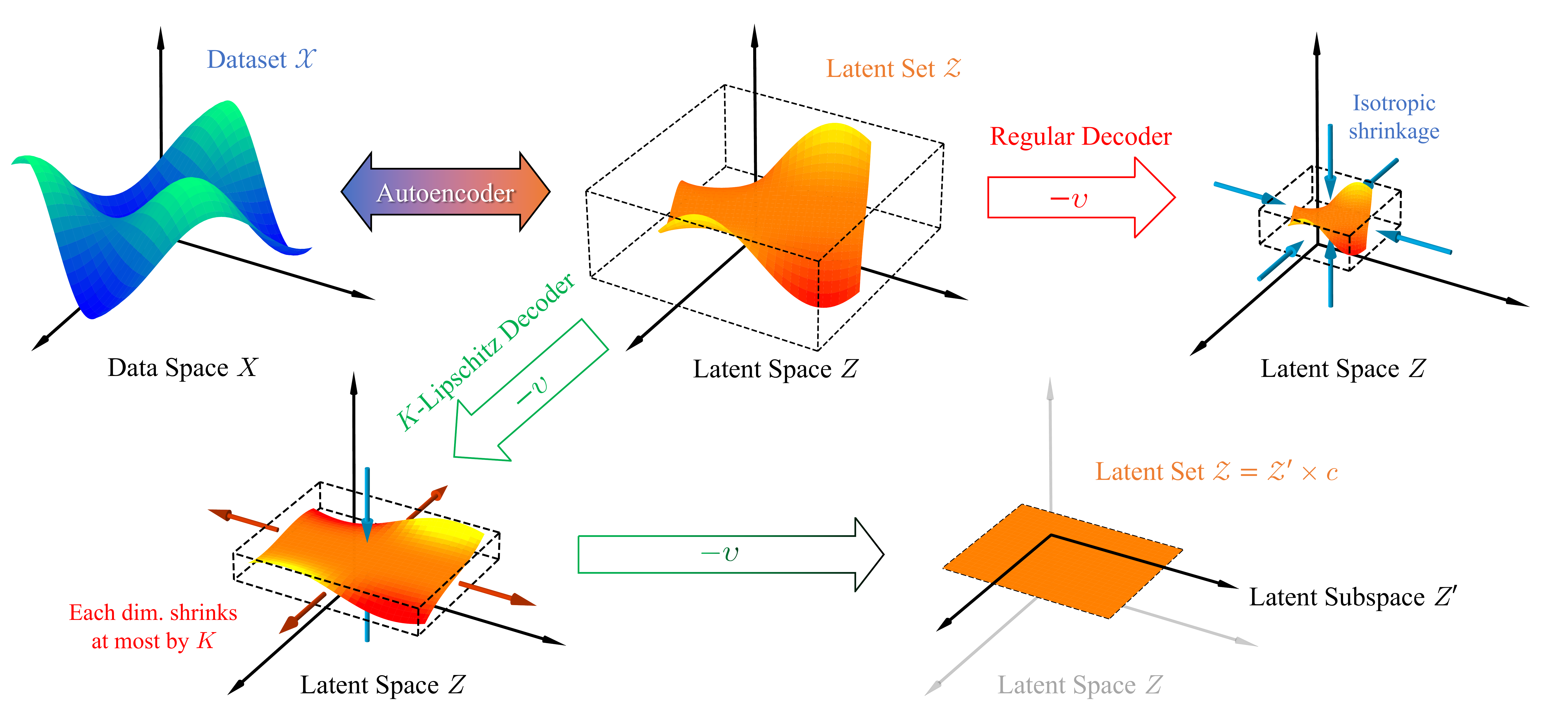}
    \caption{Flattening the latent set via Least Volume.}
    \label{fig:least_volume}
\end{figure}

\emph{Least Volume (LV)}~\cite{qiuyi2024compressing} is the method that can help us obtain such an autoencoder with a least dimensional latent space $Z$. Its key intuition is that a flat piece of paper consumes less space than when it curls. Likewise, among all the latent sets homeomorphic to the dataset $\calX$, a curved latent set $\calZ = g(\calX) \subset Z$ can only be enclosed by a cuboid of much larger volume than a cuboid that encloses its flattened counterpart (as shown in Fig.~\ref{fig:least_volume}). 
Suppose in the following \emph{conceptual} procedure we are given an autoencoder $(e_{\btheta}, g_{\btheta})$ of great complexity such that both $e_{\btheta}:X\to Z$ and $g_{\btheta}:Z\to X$ can approximate any continuous function. We keep the reconstruction loss (\ref{eq:rec}) minimized so that $\calZ$ remains homeomorphic to $\calX$, while additionally morphing $\calZ$'s geometry by adjusting $\btheta$.
We denote the standard deviation of each latent space dimension by
\begin{equation}
    \sigma_{\calZ, i} := \sqrt{\textrm{var}(\calZ)_i},
\end{equation}
and make the cuboid's side lengths equal to $\sigma_{\calZ, i}$, so that the cuboid has \emph{volume} $L_{\calZ} := \prod_{i=1}^{\dim Z}\sigma_{\calZ, i}$. Now if we reduce $L_\calZ$, it reaches zero only when at least one $\sigma_{\calZ, i}$ vanishes, indicating that $\mathcal{Z}$ is flattened and compressed into a linear subspace, \ie, $\calZ=\calZ'_1\times c_1 \subset Z_1'\times c_1 \subset Z$ where $c_1$ is a constant. 
Naturally, the way to proceed would be to extract $Z'_1$ as the new latent space of lower dimension, try compressing $\calZ'_1\subset Z'_1$ inside into a new subspace $Z'_2\times c_2\subset Z'_1$ again (\ie, $\calZ'_1 = \calZ'_2\times c_2 \subset Z'_2\times c_2\subset Z'_1$) by minimizing $L_{\calZ'_1}$, and repeat the same extraction and compression recursively for $n$ times until $\calZ'_n$ cannot be flattened any more: that is, its volume $L_{\calZ_n'}$ cannot reach zero. 
The final $Z'_n$ is then the least dimensional latent space $Z'$ that we seek, which can be extracted by the \emph{projection} $\pi: Z'\times c\to Z'$ where $c\coloneqq \bigtimes_{i=1}^n c_i$ is the cartesian product. Here $\pi$ is continuous and in fact is also a homeomorphism~\cite{qiuyi2024compressing}. It follows that $(\pi\circ e_{\btheta},\, g_{\btheta}\circ\pi^{-1})$ is the desired continuous autoencoder that maps $\calX$ to the least dimensional latent space $Z'$, in which $\calZ'=\pi\circ e_{\btheta}(\calX)\subset Z'$ is still homeomorphic to $\calX$.

In practice this conceptual recursive procedure is difficult to implement, so in~\cite{qiuyi2024compressing} the authors further introduced an increment $\eta > 0$ and instead minimize the geometric mean
\begin{equation}
L_{\calZ, \eta} = \sqrt[\dim Z]{\prod_{i=1}^{\dim Z} (\sigma_{\calZ,i} + \eta)}
\end{equation}
in order to conveniently achieve a similar outcome, while avoiding vanishing gradient issue when $\sigma_{\calZ,i}$ goes to zero or exploding gradient issue when $\dim Z$ is large. In this case, optimization can be carried all the way through with no recursion as it allows the standard deviation across more than one dimension to reach zero simultaneously ($\sigma_{\calZ, i}\simeq 0$ for multiple $i$'s).
Nevertheless, reducing $L_{\calZ, \eta}$ with no further constraints will trivially drive all $\sigma_{\mathcal{Z}, i}$ close to zero without learning anything useful, because $\mathcal{Z}$ may shrink \emph{isotropically} to na\"ively shorten all $\sigma_{\mathcal{Z}, i}$ without flattening itself. To avoid this, the decoder $g_{\btheta}$ is constrained to be $K$-Lipschitz through \emph{spectral normalization}~\cite{miyato2018spectral, gouk2021regularisation} for a small $K$.  By doing so, any latent dimension $i$ with $\sigma_{\calZ,i}\simeq 0$ must also have close-to-zero variation in the original data space $X$ and thus can only correspond to a trivial data dimension, whereas a principal data dimension must retain a large variation in the latent space to avoid violating the Lipschitz constraint. Hence the isotropic shrinkage is prevented.
Figure~\ref{fig:least_volume} illustrates this intuition.

Thereafter, one may safely \emph{prune} these latent dimensions $i$ with $\sigma_{\calZ, i}\simeq 0$ in ascending order of $\sigma_{\calZ, i}$, until the reconstruction loss $J(\btheta)$ increases to a certain user-defined threshold $\delta$. 
By pruning dimension $i$ we mean replacing that dimension's code $z^{(j)}_i = e_{\btheta}(x^{(j)})_i$ of every sample $x^{(j)}$ with its mean $\bar{z}_i=\mathbb{E}_{x\sim p_\calX(x)}[e_{\btheta}(x)_i]$.  
It is safe to do so because the $\ell_2$ reconstruction error of the autoencoder with a $K$-Lipschitz decoder increases by at most $K \sqrt{\sum_{i\in P} \sigma_{\calZ, i}^2}$, where $P$ is the set of indices of pruned dimensions~\cite{qiuyi2024compressing}.
After pruning, $\calZ$ then has the form $\calZ = \calZ' \times \bigtimes_{i\in P} \bar{z}_i \subset Z' \times \bigtimes_{i\in P} \bar{z}_i$ and can be fed to $\pi: Z'\times \bigtimes_{i\in P} \bar{z}_i\to Z'$ to extract $\calZ'$. If we denote the composition of the encoder with this pruning operation by $\bar{e}_{\btheta, \delta}$, then the autoencoder becomes $(\pi\circ \bar{e}_{\btheta, \delta},\, g_{\btheta}\circ\pi^{-1})$.

\subsubsection{Loss function}
\label{sec:loss_function}
In summary, the \emph{Least Volume Autoencoder (LVAE)} that we want to obtain is the $(\pi\circ \bar{e}_{\btheta, \delta},\, g_{\btheta}\circ\pi^{-1})$ whose parameter $\btheta$ is the solution to the following constrained optimization problem:
\begin{eqnarray}
\label{eq:LVAE_opt}
\begin{array}{cl}
    \arg\min_{\btheta} L_{\calZ,\eta}(\btheta), & L_{\calZ, \eta}(\btheta) := L_{e_{\btheta}(\calX), \eta} \\ 
    \textrm{s.t.} & \calZ \in \left\{\calZ = e_{\btheta}(\calX) \vert  \btheta \textrm{ minimizes } J(\btheta) \right\} ,\\
    & \vert\vert g_{\btheta}(z_1) - g_{\btheta}(z_2) \vert\vert \leq K\vert\vert z_1 -z_2\vert\vert, \forall z_1, z_2 \in Z.
\end{array}
\end{eqnarray}
Yet enforcing the reconstruction constraint is hard for nonlinear autoencoders. So in practice, we resort to a simpler optimization problem: that is, to minimize the objective function 
\begin{eqnarray}
    \ell(\btheta) := J(\btheta) + \lambda L_{\calZ, \eta}(\btheta),
\label{eq:obj}
\end{eqnarray}
where $\lambda$ needs to be fine-tuned to make $J(\btheta) < \delta$. The reconstruction loss $J$ throughout this work is set to be $J(\btheta) = \E{x \sim p_{\calX}(x)}{\vert\vert g_{\btheta} \circ e_{\btheta} (x) - x \vert\vert}$.

Many of LVAE's nice properties have been shown in~\cite{qiuyi2024compressing}. In particular, when the encoder and decoder are represented by linear functions $e(x) = Ax + a$ and $g(z) = Bz + b$, with $g$ being a $1$-Lipschitz and $\textrm{rank cov}(\calX) \geq \textrm{dim} Z$, the LVAE is equivalent to Principal Component Analysis (PCA)~\cite{pearson} and recovers the principal components of $\calX$. Empirircally, the PCA-like importance ordering effect persists even when $e$ and $g$ are nonlinear, such that we can assess the importance of the latent dimensions $i$ with their standard deviations $\sigma_{\calZ, i}$. We shall leverage this property later in the experiment section.

\subsection{Conditional Sinkhorn Generative Adversarial Networks}
\label{sec:csgan}
Next, we give a brief review to the \emph{conditional Sinkhorn generative adversarial networks} (CSGANs) that we are going to use to solve the inverse problems by approximating their posterior distributions on a data-driven basis, based on the formulation introduced in~\cite{genevay2018learning, feydy2019interpolating}. 

\subsubsection{Pushforward Generative Models}
\label{sec:pushforward}

\emph{Generative adversarial networks} (GANs)~\cite{goodfellow2014generative, arjovsky2017wasserstein, gulrajani2017improved} is a subfamily of a larger family of generative models dubbed \emph{pushforward models}~\cite{cornish2020relaxing, salmona2022can, brown2022verifying} that approximate a target probability measure $\mathbb{P}_r$ with a \emph{pushforward measure} $\mathbb{P}_g$ after minimizing a certain \emph{statistical divergence} (\emph{or distance})  $D(\mathbb{P}_g, \mathbb{P}_r)$ between them. The pushforward measure $\mathbb{P}_g$ is created by transforming a probability measure $\mathbb{P}_u$\textemdash usually referred to as the \emph{latent distribution}\textemdash in a latent space $U$ through a measurable \emph{generator} function $g: U\to X$ (typically modeled with a neural network) into the data space $X$, such that 
\begin{equation}
    \mathbb{P}_g (A) = g_*\mathbb{P}_u(A) \coloneqq \mathbb{P}_{u}(g^{-1}(A))
    \label{eq_pushforward}
\end{equation}
for any $A\subset X$, $B\subset U$ that are $\mathbb{P}_g$- and $\mathbb{P}_u$-measurable sets respectively such that $A = g(B)$.
For the ease of sampling $x\sim\mathbb{P}_g$\textemdash \ie, first sampling $u\sim \mathbb{P}_u$ then producing $x=g(u)$\textemdash the latent distribution $\mathbb{P}_u$ is often a simple probability measure, \eg, a Gaussian or uniform distribution. Thanks to $g$'s high complexity~\cite{cybenko1989approximation}, $\mathbb{P}_g$ can approximate many intricate $\mathbb{P}_r$ in $X$. 

Different choices of the statistical divergence $D$ lead to different types of pushforward models. 
Among the most common, variational autoencoders~\cite{kingma2013auto, sohn2015learning} and normalizing flows~\cite{dinh2014nice, dinh2016density, kingma2018glow, kobyzev2020normalizing} maximize the log likelihood of the data samples, which is equivalent to minimizing the \emph{Kullback–Leibler divergence} $D_{KL}(\mathbb{P}_r\| \mathbb{P}_g)$. On the other hand, GANs mostly minimize statistical divergences other than $D_{KL}$, such as the \emph{Jensen-Shannon divergence} for vanilla GANs~\cite{goodfellow2014generative}, the $f$-divergence for $f$-GANs~\cite{nowozin2016f} and the \emph{Wasserstein distance} $\mathrm{OT} (\mathbb{P}_{r}, \mathbb{P}_{g})$ for Wasserstein GANs~\cite{arjovsky2017wasserstein, gulrajani2017improved}. More detailed reviews on the use of different loss function has been conducted in~\cite{hong, yi_walia, jabbar}.

\subsubsection{Sinkhorn GANs}
\label{sec:sinkhorn_gans}

The Wassertein distance is based on optimal transport theory and is defined as 
\begin{equation}
    \mathrm{OT} (\mathbb{P}_{r}, \mathbb{P}_{g}) \coloneqq
    \min_{\mathbb{P}_{X, \hat X} \in \Pi(\mathbb{P}_{r}, \mathbb{P}_{g})}
    \mathbb{E}_{{x, \hat x}\sim\mathbb{P}_{X, \hat X}}[c({x,\hat x})]
\end{equation}
where $\Pi(\mathbb{P}_{r}, \mathbb{P}_{g})$ is the family of joint distributions $\mathbb{P}_{X, \hat X}$ whose
marginal distributions equal $\mathbb{P}_{r}$ and $\mathbb{P}_{g}$ respectively, and $c$ is usually a symmetric cost function, such as $\ell_1$ or $\ell_2$ distance. Due to the computational challenges associated with the numerical evaluation of $\mathrm{OT}$, many researchers have turned their attention to its regularized variations, such as the ones including an \emph{entropy regularization} term with weight $\rho$, written as~\cite{cuturi2013sinkhorn, feydy2019interpolating, peyre2019computational}:
\begin{equation}
    \mathrm{OT}_\rho (\mathbb{P}_{r}, \mathbb{P}_{g}) \coloneqq
    \min_{\mathbb{P}_{X, \hat X} \in \Pi(\mathbb{P}_{r}, \mathbb{P}_{g})}
    \mathbb{E}_{{x, \hat x}\sim\mathbb{P}_{X, \hat X}}[c({x,\hat x})] \\
    + \rho D_{KL}(\mathbb{P}_{X, \hat X} \| \mathbb{P}_{r} \times \mathbb{P}_{g}).
    \label{eq_primal}
\end{equation}
The above expression possesses strong duality and can thus be efficiently evaluated after writing its concave dual formulation:
\begin{align}
    \mathrm{OT}_\rho (\mathbb{P}_{r}, \mathbb{P}_{g}) 
    &= 
    \max_{\alpha,\,\beta}
    \mathbb{E}_{{x}\sim \mathbb{P}_{r}}[\alpha(x)] + \mathbb{E}_{{\hat x}\sim \mathbb{P}_{g}}[\beta({\hat x})] \nonumber\\
    &\quad -
    \rho \mathbb{E}_{x,\,{\hat x}\sim \mathbb{P}_{r}\times \mathbb{P}_{g}} 
    \left[\exp(v({x,\hat x}) / \rho)-1\right],
    \label{eq_dual} \\[10pt]
    v({x, \hat x}) 
    &\coloneqq \alpha(x) + \beta({\hat x}) - c({x,\hat x}).
    \label{eq_veval}
\end{align}
Here $\alpha$ and $\beta$ are the Lagrange multipliers that can be rapidly retrieved through the Sinkhorn–Knopp algorithm~\cite{cuturi2013sinkhorn, feydy2019interpolating} based on coordinate ascent. 

However, the above regularization introduces a bias, since for $\mathbb{P}_r = \mathbb{P}_g$ we no longer have  $\mathrm{OT}_\rho (\mathbb{P}_r, \mathbb{P}_g) =  0$~\cite{feydy2019interpolating}. To mitigate this bias, Genevay~{\it et~al.} further introduced the \emph{Sinkhorn GAN}\footnote{Unlike the vanilla GAN, there is actually nothing \emph{conventionally} ``adversarial'' in this pushforward model's training process, since we do not need to train any discriminator to compete with the generator. Regardless, the discriminator's training is now replaced by the Sinkhorn–Knopp algorithm evaluating the Sinkhorn divergence, which may also be regarded as something ``adversarial'' in a broader sense since it ``trains'' the Lagrangian multipliers $\alpha$ and $\beta$, so we still identify this model as a type of GAN for convenience. }~\cite{genevay2018learning} 
that instead minimizes the \emph{Sinkhorn divergence} $\mathrm{S}_\rho(\mathbb{P}_r,  \mathbb{P}_g)$, which is a symmetric variation of $\mathrm{OT}_\rho$ defined as~\cite{feydy2019interpolating, genevay2018learning, peyre2019computational}:
\begin{equation}
    \mathrm{S}_\rho (\mathbb{P}_{r}, \mathbb{P}_{g}) 
    =
    \mathrm{OT}_\rho (\mathbb{P}_{r}, \mathbb{P}_{g})
    - \frac{1}{2} \mathrm{OT}_\rho (\mathbb{P}_{r}, \mathbb{P}_{r})
    - \frac{1}{2} \mathrm{OT}_\rho (\mathbb{P}_{g}, \mathbb{P}_{g}).
    \label{eq_sinkhorn}
\end{equation}
This distance stands out from the rest for its stability and fast evaluation that spares the costly and unstable training of a discriminator function (as in~\cite{goodfellow2014generative, arjovsky2017wasserstein}), and its ability to interpolate between $\mathrm{OT}$ and the \emph{maximum mean discrepancy} (MMD)~\cite{smola2007hilbert} by adjusting $\lambda$~\cite{feydy2019interpolating}. 

\subsubsection{Conditional Sinkhorn GANs}
\label{sec:csgan_sub}
Suppose along with the dataset $\calX$ following the distribution $\mathbb{P}_r^X$, we are also provided with the \emph{condition set} $\calY$ endowed with the distribution $\mathbb{P}_r^Y$. Very often $\calX$ is dependent on $\calY$ and we are interested in learning their relationship through their joint distribution $\mathbb{P}_r^{X, Y}$. To this end, all pushforward models introduced above can be generalized to approximate the conditional distributions $\mathbb{P}_r^{X\mid Y}$\textemdash as opposed to the marginal densities $\mathbb{P}_r^{X}$\textemdash by simply augmenting the generator input such that $g: U\times Y \to X$. 
This implies that the corresponding pushforward measure $\mathbb{P}_g^{X} = g_*\mathbb{P}_u$ becomes a conditional one $\mathbb{P}_g^{X\mid Y=y} = g[y]_*\mathbb{P}_u$, where $g[y]$ denotes the restriction $g|_{U\times y}:U\to X$ for $y\in\calY\subset Y$. Consequently, training $g$ now becomes enforcing $\mathbb{P}_g^{X\mid Y}$ to approximate the target conditional distribution $\mathbb{P}_r^{X\mid Y}$ by minimizing the divergence $D(\mathbb{P}_r^{X, Y}, \mathbb{P}_g^{X, Y})$ between the target joint distribution $\mathbb{P}_r^{X, Y} = \mathbb{P}_r^{X\mid Y}\mathbb{P}_r^{Y}$ and the pushforward joint distribution $\mathbb{P}_g^{X, Y}=\mathbb{P}_g^{X\mid Y}\mathbb{P}_r^{Y}$. 

We will refer to the resulting conditional variation of Sinkhorn GANs\textemdash for which $D=\mathrm{S}_\rho$\textemdash as the \emph{conditional Sinkhorn GANs} (CSGANs). 
In practice, the Sinkhorn divergence $\mathrm{S}_\rho(\mathbb{P}_r^{X, Y}, \mathbb{P}_g^{X, Y})$ can be conveniently evaluated by sampling batches of data points $(x^{(j)}, y^{(j)})\sim \mathbb{P}_r^{X, Y}$ and $(\hat{x}^{(j)}, y^{(j)})\sim \mathbb{P}_g^{X, Y}$ from the two joint distributions respectively and passing them to the Sinkhorn algorithm. $\mathbb{P}_r^{X, Y}$ can be sampled easily by picking $(x^{(j)}, y^{(j)})$ straight out of the digital dataset $\calD = \{(x^{(j)}, y^{(j)})\}_n$ uniformly, whereas $\mathbb{P}_g^{X, Y}$ can be sampled by taking $y^{(j)}$ out of $\calD$ uniformly (thus $y^{(j)}\sim\mathbb{P}_r^Y$) and passing them along with $u^{(j)}\sim \mathbb{P}_u$ to $g$ to produce the predictions $\hat{x}^{(j)} = g(u^{(j)}, y^{(j)})$ and form the tuples $(\hat{x}^{(j)}, y^{(j)})$. More technical details about the implementation of the Sinkhorn algorithm can be found in~\cite{feydy2019interpolating, feydy2019geometric}.

\subsection{CSGANs in Least Volume Latent Spaces}
\label{sec:lvae_csgan}

Although the CSGANs can approximate the conditional distributions $\mathbb{P}^{X\mid Y}_r$, when the data space $X$ and the condition space $Y$ has very high dimensionality, the CSGAN's training can decelerate significantly, because evaluating the Sinkhorn divergence $\mathrm{S}_\rho(\mathbb{P}_r^{X, Y}, \mathbb{P}_g^{X, Y})$ typically requires the calculation of $c\left((x^{(j)}, y^{(j)}), (\hat{x}^{(k)}, y^{(k)})\right)$ in (\ref{eq_veval}) over $n\geq 1000$ pairs of $(j, k)$, and the evaluation of $c$ among all pairs normally has time complexity $O\left(n(\dim X + \dim Y)\right)$ (\eg, when $c$ is the $\ell_1$ or $\ell_2$ distance). 
In addition, it is crucial to find out a reasonable latent space dimension $\dim Z$ for the CSGAN's generator $g:U\times Y\to X$. If $\dim Z$ is lower than $\dim \mathcal{X}$, then due to the \emph{rank theorem}~\cite{lee2012smooth}, the generator $g$'s image $g(\mathcal{Z})$ can at best only be a small subset of measure zero in $\mathcal{X}$ and unable to generate all points on $\mathcal{X}$. Consequently, a distribution supported by $\mathcal{X}$ cannot be approximated well by the resulting pushforward measure $\mathbb{P}_g^{X\mid Y=y}$~\cite{chen2022learning}. Moreover, the difference between $\dim \calX$ and $\dim \calY$ may provide us some insights into the relative information content between $X$ and $Y$, allowing us to anticipate the uncertainty of $X$ even prior to obtaining $\mathbb{P}^{X\mid Y}_r$, so retrieving them could be beneficial. This hypothesis will be detailed and verified in the later experiments in \S\ref{sec:oilextraction}.

To overcome these issues, we can integrate the LVAEs with the CSGANs to form the \emph{latent CSGANs}. 
Specifically, suppose we obtained two trained LVAEs $(e_x:X\to Z_x, d_x:Z_x\to X)$ and $(e_y:Y\to Z_y, d_y: Z_y\to Y)$ for the dataset $\calX$ and the condition set $\calY$ respectively, whose latent spaces $Z_x$ and $Z_y$ have their dimensions reduced as much as possible. 
We first transform the dataset $\calX\subset X$ into the latent set $\mathcal{Z}_x = e_x(\calX)\subset Z_x$, and the condition set $\calY\subset Y$ into the latent set $\mathcal{Z}_y = e_y(\calY)\subset Z_y$. 
Thus, the latent codes $z_x\in\calZ_x$ and $z_y\in\calZ_y$ jointly follow the distribution $\mathbb{P}_e^{Z_x,Z_y} \coloneqq (e_x\times e_y)_*\mathbb{P}_r^{X,Y}$. 
After these encodings, we can train the latent CSGAN with the latent generator $g: U\times Z_y\to Z_x$ in $Z_x\times Z_y$ similarly to \S\ref{sec:csgan_sub}: that is, minimizing $\mathrm{S}_\rho(\mathbb{P}_g^{Z_x, Z_y}, \mathbb{P}_e^{Z_x, Z_y})$ in order to make $\mathbb{P}_g^{Z_x\mid Z_y=z_y}\coloneqq g[z_y]_*\mathbb{P}_u\simeq \mathbb{P}_e^{Z_x|Z_y=z_y}$.
This enables $g[z_y]$ to approximately produce $z_x\in\mathcal{Z}_x\sim \mathbb{P}_e^{Z_x|Z_y=z_y}$.
With this latent setup, the latent generator $g: U\times Z_y\to Z_x$ is more efficient to train with the Sinkhorn algorithm especially when $\dim Z_x \ll \dim X$ or $\dim Z_y \ll \dim Y$, and we can set $\dim U = \dim Z_x$ to prevent dimension mismatch since $\dim Z_x\geq \dim \mathcal{X}$. Apart from this, $\dim Z_x$ and $\dim Z_y$ also estimate $\dim \calX$ and $\dim \calY$ by upper-bounding them.

Finally, in order to sample $x\sim \mathbb{P}_r^{X\mid Y=y}$ given $y\in\calY$, we can first obtain $z_y = e_y(y)$, then feed it along with $u\sim \mathbb{P}_u$ into $g$ to produce $z_x = g(u, z_y)\sim \mathbb{P}_g^{Z_x|Z_y=z_y}$, and at last transform $z_x$ back into $X$ through $x = d_x(z_x)$, so that the generated $x = d_x\circ g(u, e_y(y))$ comes from the conditional pushforward measure $\mathbb{P}_g^{X\mid Y=y} 
\coloneqq (d_x\circ g[e_y(y)])_*\mathbb{P}_u$. 
Despite the latent training process, $\mathbb{P}_g^{X\mid Y}$ still approximates $\mathbb{P}_r^{X\mid Y}$, given that
\begin{eqnarray}
\begin{array}{cl}
\mathbb{P}_g^{X\mid Y=y}
&= d_{x*} \left(g[e_y(y)]_*\mathbb{P}_u\right)
= d_{x*} \left(\mathbb{P}_g^{Z_x\mid Z_y = e_y(y)}\right) \\
&\simeq 
d_{x*} \left(\mathbb{P}_e^{Z_x\mid Z_y = e_y(y)}\right)
= d_{x*} \left(e_{x*}\mathbb{P}_r^{X\mid Y = y}\right) \\
&= (d_x\circ e_x)_* \mathbb{P}_r^{X\mid Y = y}
=\mathbb{P}_r^{X\mid Y=y}.
\end{array}
\end{eqnarray}


\section{Results and Discussion}
\label{sec:experiment}

In this section, we verify the latent CSGAN framework's ability to solve high dimensional inverse problems in a Bayesian manner, and analyze these results to provide further insights. First, we apply it to the relatively simple Kraichnan-Orszag ODEs to identify the system's bimodal source, and illustrate the latent CSGAN's low dimensional predictions to uncover its strength and weakness visually. Next, we extend the latent CSGANs to the more complicated higher-dimensional two-phase flow problem of multi-well oil extraction. Apart from studying the latent CSGAN's accuracy in approximating the target posterior distribution, we demonstrate that the dimension of the least volume latent space can serve as an accurate estimator of the intrinsic dimension of the dataset\textemdash even when the dataset is nonlinear\textemdash and further reveal that the difference between the dimensions of the posterior's input condition and output may help predict the uncertainty in the target posterior beforehand, assisting the inverse problem's optimal configuration in future projects. 

\subsection{Source Identification in Kraichnan-Orszag Three-mode Problem}
\label{sec:ko_ode}

The Kraichnan-Orszag (KO) three model problem is a set of nonlinear ordinary differential equations (ODE) of particular interest and has been previously investigated by several authors~\cite{orszag, wan-karniadakis, ma-zabaras, bilionis_mogp}. It is described by the system of ODEs:

\begin{eqnarray}
\label{eq:KO_ode}
    \begin{array}{ccl}
        \displaystyle{\frac{dy_1}{dt}} & = & y_1y_3, \\
        \displaystyle{\frac{dy_2}{dt}} & = & -y_2 y_3, \\
        \displaystyle{\frac{dy_3}{dt}}
        & = & -y_1^2 + y_2^2
    \end{array}
\end{eqnarray}
subject to initial random conditions at $t=0$. The stochastic initial conditions are defined by:
\begin{eqnarray}
\begin{array}{ccc}
    y_1(0) = 1 & y_2(0) = \xi_1 & y_3(0) = \xi_2, \\
\end{array}
\end{eqnarray}
where
\begin{eqnarray}
\label{eq:xi_prior}
    \xi_1 \sim \calU[-0.1, 0.1],\quad\xi_2 \sim \calU[-1, 1].
\end{eqnarray}
The solution $y(t; \bxi)$ of (\ref{eq:KO_ode}) is discontinuous with respect to $\bxi$ at $\xi_1 = 0$. 

In the inverse problem considered here, we are interested in inferring the initial conditions $y_2(0)$, $y_3(0)$ (or equivalently the posterior distributions of $\bxi$) by observing only the behavior of $y_1(t)$ in the time interval $t\in [0, 30]$. In other words, we aim to retrieve the posterior $p(\bxi\mid y_1)$. In order to generate synthetic data to solve this problem on a data-driven basis, we solve numerically the ODE using a 4th order Runge-Kutta method with a time step $\Delta t=0.003$, resulting in a response vector $\by = [y_1(\bt;\bxi), y_2(\bt; \bxi), y_3(\bt; \bxi)] \in \R^{3\times 10^4}$. We run $N = 2048$ numerical simulations of the ODE that correspond to an equal number of Monte Carlo samples $\bxi$ drawn from the prior (\ref{eq:xi_prior}), to be used as the training data for the CSGAN.

\subsubsection{Model Configurations and Training}
\begin{figure}
     \centering
     \begin{subfigure}[b]{0.45\textwidth}
         \centering
         \includegraphics[width=\textwidth]{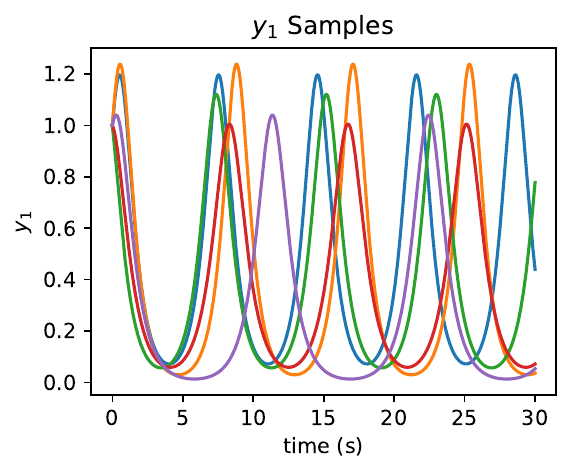}
         \caption{5 samples of $y_1(t)$}
         \label{fig:ko_y1}
     \end{subfigure}
     \hfill
     \begin{subfigure}[b]{0.45\textwidth}
         \centering
         \includegraphics[width=\textwidth]{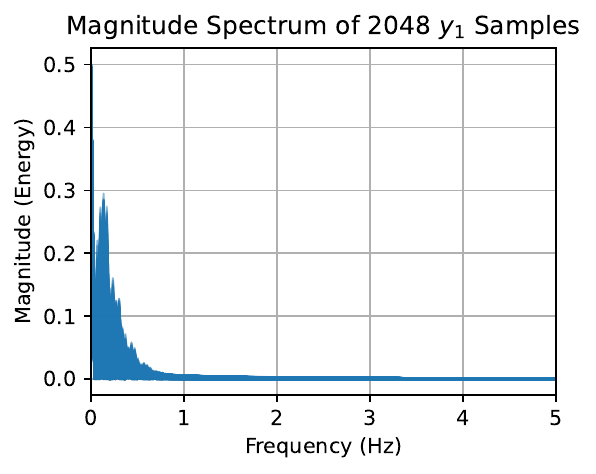}
         \caption{Magnitude spectrum of $y_1(t)$ curves}
         \label{fig:ko_spectrum}
     \end{subfigure}
        \caption{FFT result of $y_1(t)$ samples}
        \label{fig:ko_dataset}
\end{figure}

We start by preprocessing the dataset. Each of the 2048 input conditions $y_1(t)$ consists of 10001 dimensions coming from the $\Delta t = 0.003$ sampling spacing over the $[0, 30]$ time span. Since FFT shows that $y_1(t)$ in general has bandwidth $\leq$ 4 Hz (Fig.~\ref{fig:ko_dataset}), for the ease of the following tasks, we linearly interpolate and uniformly resample $y_1(t)$ into 256-D vectors at the sampling rate 8.5 Hz, so as to preliminarily reduce $y_1(t)$'s dimension by leveraging the \emph{Nyquist–Shannon sampling theorem}~\cite{shannon1949communication}. Thereafter, we normalize both $\bxi$ and $y_1(t)$ to the range $[0,1]$, as a normal practice to facilitate the training of neural networks. We reserve 512 samples as the test samples and keep the rest 1536 samples as the training set to train the latent CSGAN as follows. 

To approximate the target posterior $p(\bxi\mid y_1)$ efficiently, first we use LVAE to further reduce the dimension of the 256-D $y_1(t)$. We model the encoder with an 1D convolutional neural network and the decoder with an 1D deconvolutional neural network. Their architectures and hyperparameters are listed in Table~\ref{tab:ko_lvae}. The dimension of the autoencoder's latent space is set to 50. We set the parameter $\eta$ in the loss function (\ref{eq:obj}) to 0.004, and to prevent the LVAE from getting trapped at local minima when $\lambda$ is too large at the start, we gradually increase $\lambda$ from 0 to 0.2 at constant speed throughout the training period of 2000 epochs. We set the batch size to 50, so each of these 2000 epochs contains $1536/50\simeq 31$ iterations. We use Adam~\cite{kingma2014adam} with a learning rate of 0.0001 for optimization.

After the LVAE finishes compressing $y_1(t)$ to $n$ certain dimensions, we extract this low dimensional latent subspace and train a latent CSGAN with the generator architecture in Table~\ref{tab:ko_csgan}. The generator's output dimension is equal to $\dim\bxi=2$, and for its two inputs\textemdash latent noise $U$ and condition $Y$\textemdash we set $\dim U=\dim\bxi=2$ and $\dim Y = n \,(=7)$. The latent distribution $\mathbb{P}_u$ over $U$ is set to the uniform distribution $\calU[0, 1]^{2}$. We make $\rho = 0.05$ for the evaluation of the Sinkhorn divergence $\mathrm{S}_\rho$ ($\ref{eq_sinkhorn}$). The CSGAN is trained for 2000 epochs, with the batch size 128. The optimizer is Adam with a learning rate of 0.0001.

\subsubsection{Results: Least Volume Dimension Reduction}

\begin{figure}%
  \begin{minipage}{.25\textwidth}
    \begin{subfigure}{\textwidth}
      \includegraphics[width=\textwidth]{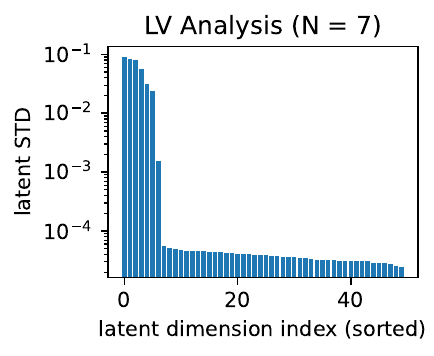}
      \caption{LVAE}
      \label{fig:ko_lv}
    \end{subfigure}
    \begin{subfigure}{\textwidth}
      \includegraphics[width=\textwidth]{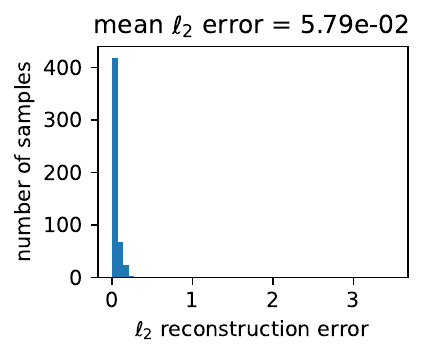}
      \caption{LVAE's $\ell_2$ Error}
      \label{fig:ko_lv_error}
    \end{subfigure}
  \end{minipage}
  \hfill
  \begin{minipage}{.25\textwidth}
    \begin{subfigure}{\textwidth}
      \includegraphics[width=\textwidth]{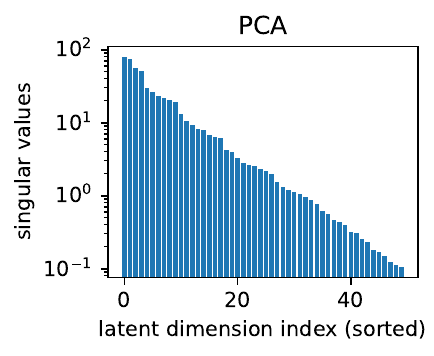}
      \caption{PCA}
      \label{fig:ko_pca}
    \end{subfigure}
    \begin{subfigure}{\textwidth}
      \includegraphics[width=\textwidth]{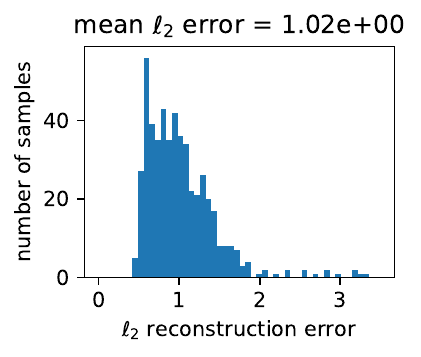}
      \caption{PCA's $\ell_2$ Error}
      \label{fig:ko_pca_error}
    \end{subfigure}
  \end{minipage}
  \hfill
  \begin{minipage}{.45\textwidth}
    \begin{subfigure}{\textwidth}
      \includegraphics[width=\textwidth]{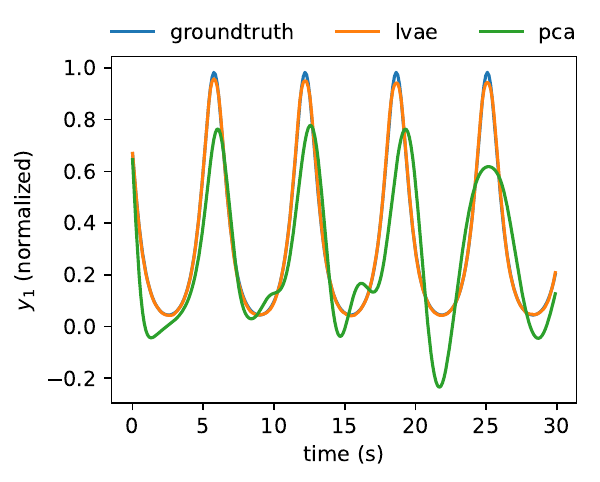}
      \caption{Example of reconstructing $y_1(t)$ from 7 latent dimensions. Since the $y_1(t)$ dataset does not live in a 7-D \emph{linear} subspace, the linear PCA has difficulty reconstructing $y_1(t)$, whereas the nonlinear model LVAE has none. 
      }
      \label{fig:ko_rec}
    \end{subfigure}
  \end{minipage}
  \caption{Dimension Reduction Result of KO Dataset}
\end{figure}

The LVAE ends up compressing $y_1(t)$ into $n=7$ dimensions from 50. This is reflected in Fig.~\ref{fig:ko_lv}, where we can see there are 7 latent dimensions whose latent STDs are orders of magnitude larger than the rest 43 dimensions. Indeed, pruning these 43 trivial latent dimensions does not bring any noticeable change to the LVAE's $\ell_2$ reconstruction error displayed in the histogram Fig.~\ref{fig:ko_lv_error}. 

Comparing the LVAE's result to PCA's can further help us conclude that the dataset of $y_1(t)$ is highly nonlinear. Figure~\ref{fig:ko_pca} shows the 50 largest singular values of PCA, where there is no apparent plummet similar to that one in Fig.~\ref{fig:ko_lv}. In addition, PCA's reconstruction error is about 20 times larger than that of the LVAE when we only keep the 7 latent dimensions of the largest singular values while discarding the rest (Fig.~\ref{fig:ko_pca_error}). This indicates that the set of $y_1(t)$ does not reside on a linear hyperplane of 7 dimensions. In fact, we can only achieve a reconstruction error comparable to LVAE's when we keep no less than 30 principal dimensions, which suggests that the $y_1(t)$ dataset should live on a curved 7-D manifold that is embedded in a 30-D linear space. For illustration, Figure~\ref{fig:ko_rec} superimposes the LVAE's and PCA's reconstructions of a given $y_1(t)$ curve from 7 principal latent dimensions. It is clear that LVAE has far superior dimension reduction and reconstruction capability.

\subsubsection{Results: Bayesian Inference of Posterior $p(\bxi\mid y_1)$}
\label{sec:ko_cgan}
\begin{figure}[bt!]
    \centering
    \includegraphics[width=\textwidth]{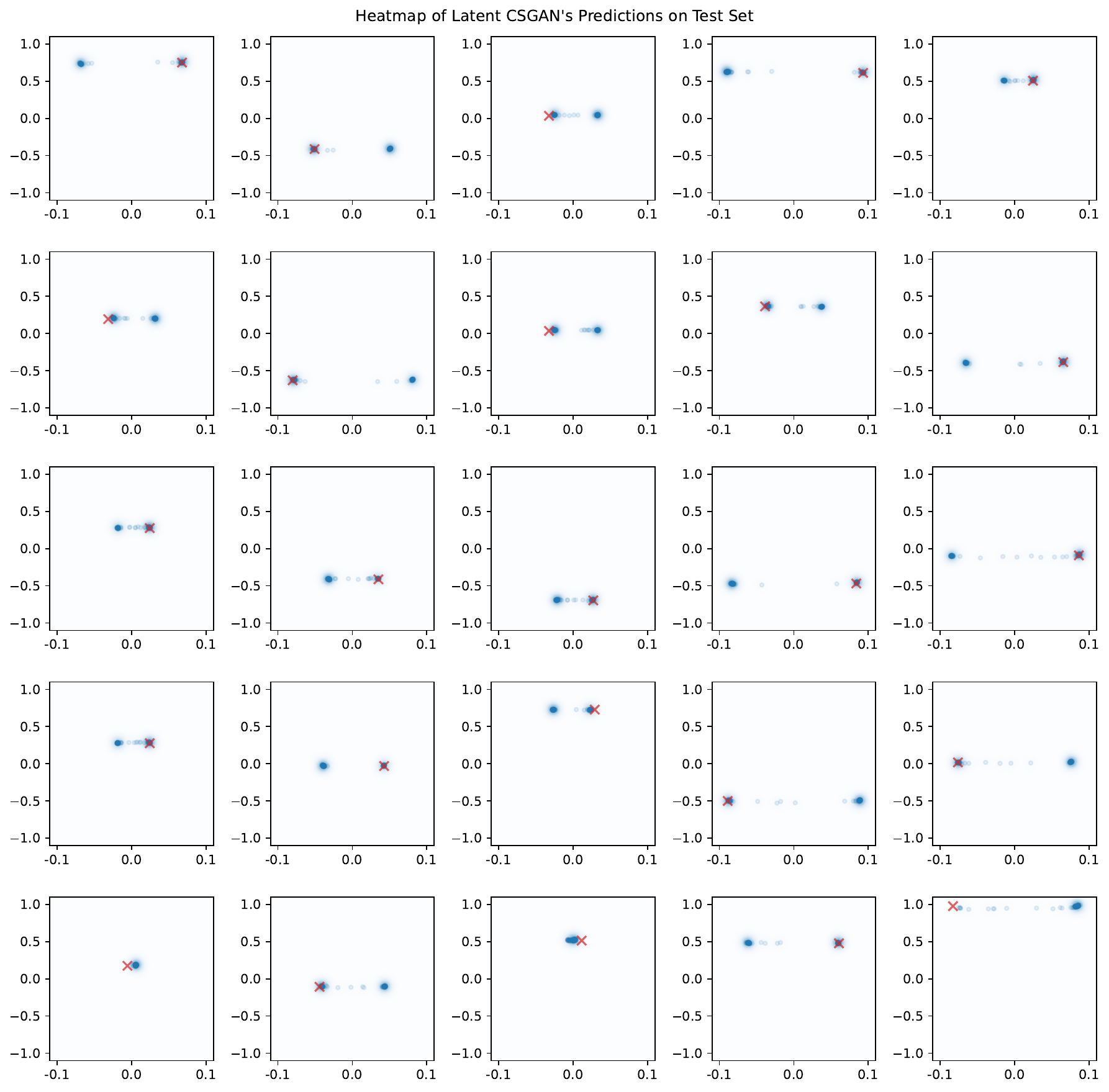}
    \caption{Heatmap of predictions made by latent CSGAN on 25 test samples. Here the red `$\times$' signs denote the groundtruth initial conditions $\bxi$ (or $y_2(0)$ and $y_3(0)$) and the scattered translucent blue dots are the posterior samples. For each case, a Gaussian filter of $\sigma=5$ is applied on the 2D histogram (on a $100\times 100$ grid over $[-0.1, 0.1]\times[-1,1]$) of these 100 predictions to produce the bluish heatmap, in order to better highlight the high density area of predictions.}
    \label{fig:ko_cgan}
\end{figure}

\begin{figure}[hbt!]
    \centering
    \begin{subfigure}[b]{0.45\textwidth}
         \centering
         \includegraphics[width=\textwidth]{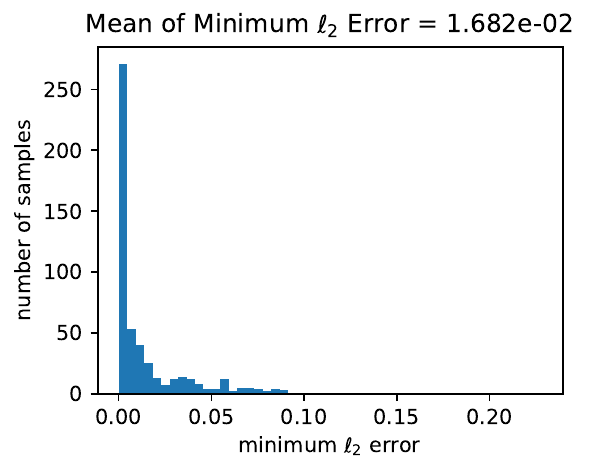}
         \caption{Quantitative distribution}
         \label{fig:ko_cgan_error}
     \end{subfigure}
     \hspace{0.1in}
    \begin{subfigure}[b]{0.43\textwidth}
         \centering
         \includegraphics[width=\textwidth]{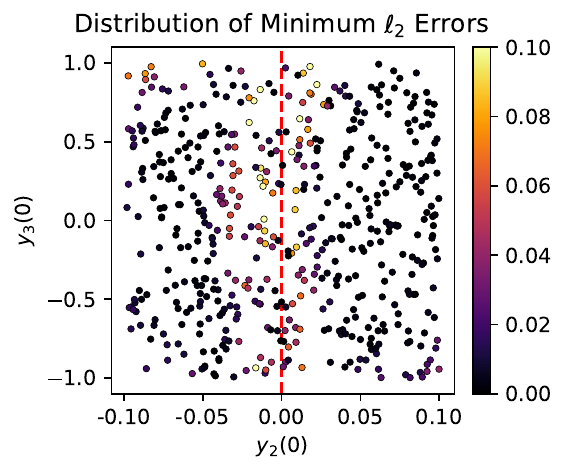}
         \caption{$\bxi$-wise spatial distribution}
         \label{fig:ko_cgan_dis}
     \end{subfigure}
     
    \caption{Distributions of Minimum $\ell_2$ Prediction Errors of Test Samples}
    \label{fig:ko_cgan_error_all}
\end{figure}

The CSGAN is then trained in the LVAE's 7-D latent space to approximate the posterior distribution $p(\bxi\mid y_1)$. To investigate the latent CSGAN's stochastic predictions' accuracy, we let the generator $g$ produce 100 predictions $\Xi^{(j)}=\left\{\hat{\bxi}\mid \hat{\bxi} = g\left(u, e(y_1^{(j)})\right), \, u\sim\mathbb{P}_u\right\}$ for each condition $y_1^{(j)}$ of the 512 test cases, then we both illustratively display the heatmap of these predictions along with their groundtruths $\bxi$ for 25 test cases in Fig.~\ref{fig:ko_cgan}, and quantitatively evaluate the minimum $\ell_2$ prediction error $\epsilon^{(j)}$ of each test case $j$\textemdash defined as $\epsilon^{(j)} = \min_{\hat{\bxi}\in\Xi^{(j)}}\|\hat{\bxi} - \bxi^{(j)}\|$\textemdash and show their distribution in the histogram Fig.~\ref{fig:ko_cgan_error}. 

The predictions in Fig.~\ref{fig:ko_cgan} clearly uncovers the bimodality of the posterior $p(\bxi\mid y_1)$ under almost all $y_1$ conditions, and these two modes appears symmetric \wrt  the axis $y_2 = 0$. This agrees well with the symmetry inherent to the KO-ODE (\ref{eq:KO_ode}): that is, for all $t$, ${dy_1}/{dt}$ and ${dy_3}/{dt}$ are independent of the sign of $y_2(t)$, whereas ${dy_2}/{dt}$ simply retains its magnitude and flips its sign as $y_2(0)$ flips its. Therefore, if $(y_1(t), y_2(t), y_3(t))$ is the solution to the ODE with initial condition $y_1(0) = 1, y_2(0)=\xi_1, y_3(0)=\xi_2$, then $(y_1(t), -y_2(t), y_3(t))$ satisfies the ODE with initial condition $y_1(0) = 1, y_2(0)=-\xi_1, y_3(0)=\xi_2$ and it is impossible to infer the sign of $y_2(0)$ by observing $y_1(t)$ alone. Thus, the bimodal structure of the posterior solution obtained by the CSGAN as shown in Fig.~\ref{fig:ko_cgan} should be expected.
The distribution of the minimum $\ell_2$ errors in Fig.~\ref{fig:ko_cgan_error} further shows that in most cases, the target $\bxi$ aligns well with either one of these two modes, as the mean error $0.0168$ is relatively marginal compared to the length scale of $\bxi$. Overall, the model captures accurately the structure of the complex posterior density, in particular near the discontinuity point $\xi_1 = 0$, which would be infeasible to achieve if we instead use a unimodal regression model to approximate the bimodal posterior or standard Bayesian inference techiques such as Markov Chain Monte Carlo. 

It should be noted that the latent CSGAN's prediction is outstanding but still not perfect. There are two flaws that can be spotted in Fig.~\ref{fig:ko_cgan}. We briefly discuss them here and leave their improvement to future studies. First, though scarce, often there are a few number of invalid predictions scattered in between these two modes (\eg, the 3rd row, 5th column of Fig.~\ref{fig:ko_cgan}). This can be attributed to the \emph{intrinsic flaw} of the pushforward generative models. Specifically, for the ease of sampling, the latent distribution $\mathbb{P}_u$ producing noise $u$ for the generator $g$ is typically a \emph{simple} distribution, in the sense that $\mathbb{P}_u$ has a support $\mathrm{supp}(\mathbb{P}_u)\subset U$ of very simple topology. For the uniform latent distribution $\mathbb{P}_u = \calU[0,1]^2$ we use, its support $[0,1]^2$ is a simply connected space, which is \emph{connected} in $U$. 
This means its image $d_x\circ g(\mathrm{supp}(\mathbb{P}_u), e_{y_1}(y_1))\subset X$ must also be connected in $X$, because both $d_x$ and $g$ are continuous neural networks, and any continuous function preserves connectedness. 
However, the closure of this image\textemdash which is also connected\textemdash is the support of the conditional pushforward measure $\mathbb{P}_g^{X\mid Y=y_1}$~\cite{brown2022verifying}.
In consequence, $\mathbb{P}_g^{X\mid Y=y_1}$ will generate points in a region connecting the two \emph{disconnected} target modes $(\xi_1, \xi_2)$ and $(-\xi_1, \xi_2)$ corresponding to $y_1$. This pathology can only be \emph{mitigated but not eliminated} if we increase $g$'s Lipschitz constant~\cite{salmona2022can}, but this entails longer training time.

The second flaw is that sometimes the CSGAN can only retrieve one of the two modes sufficiently. This seems to happen in two situations. One is when $\bxi$ is near the upper left corner of its domain $[-0.1, 0.1]\times [-1, 1]$, which can be seen in the 5th row, 5th column of Fig.~\ref{fig:ko_cgan}. This might be caused by the lack of data around that area due to the uniform sampling of $\bxi$ when preparing the dataset. Another is when $y_2(0)$ is in the vicinity of $0$, where the discontinuity of the PDE's solution happens. An example can be found in the 5th row, 1st column (or 3rd column) of Fig.~\ref{fig:ko_cgan}. It might be that in this case, $\xi_1$ and $-\xi_1$ are too close to each other while the cost function $c$ used for the Sinkhorn divergence $\mathrm{S}_\rho$ cannot efficiently differentiate between these two. Figure~\ref{fig:ko_cgan_dis} illustrates the spatial distribution of the prediction error $\epsilon^{(j)}$ in the $\bxi$ space, and the locations of high error entities agree with our discussion. However, nothing certain can be drawn at this moment, and we leave it for further investigation in future works.

\subsection{Inferring Subsurface Properties of Heterogeneous Oil Reservoirs from Observable Responses in Oil Extraction}
\label{sec:oilextraction}
Next, we move on to the more complicated inverse problem in oil production, in which we aim to infer the given heterogeneous oil reservoir's subsurface properties from its observable responses in oil extraction. The main challenge here is the high dimensionality of both the input and the output of the target posterior distribution. We will demonstrate how Least Volume can not only help approximate this high dimensional posterior by representing it in the low dimensional latent spaces, but also provide helpful information regarding the posterior's uncertainty that may help improve the inverse problem's configuration.

\subsubsection{Two-phase Flow Problem in Heterogeneous Porous Materials}

\begin{figure}[bt!]
    \centering
    \includegraphics[width=0.345\textwidth, trim={0.4cm 3.5cm 0.3cm 4cm},clip]{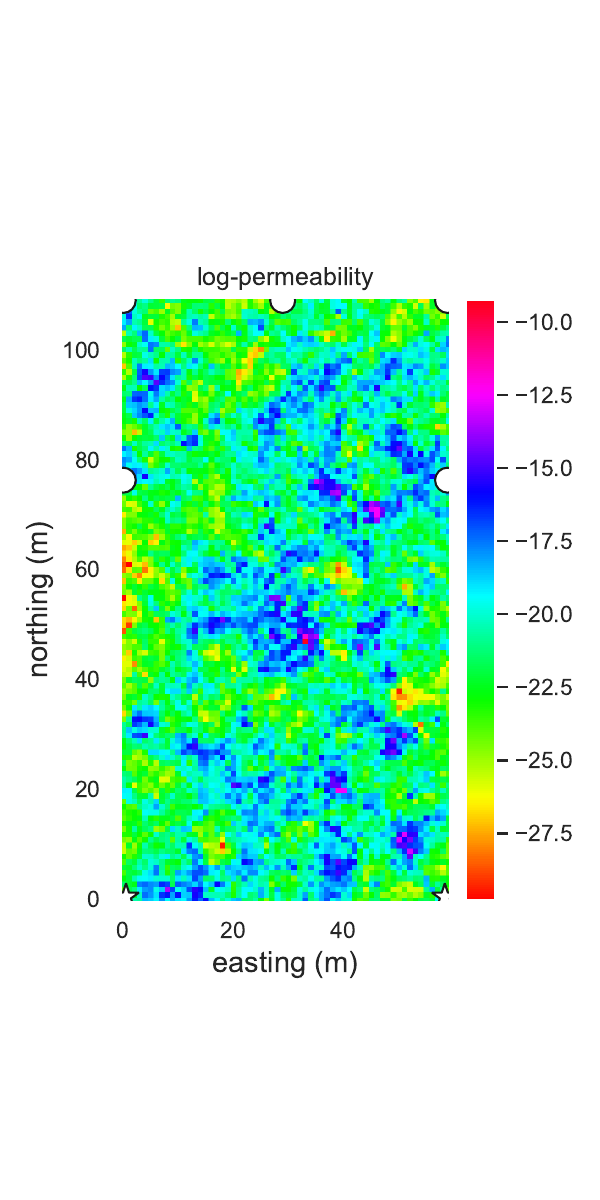}
    \hspace{1cm}
    \includegraphics[width=0.32\textwidth, trim={0.4cm 3cm 0.3cm 4cm},clip]{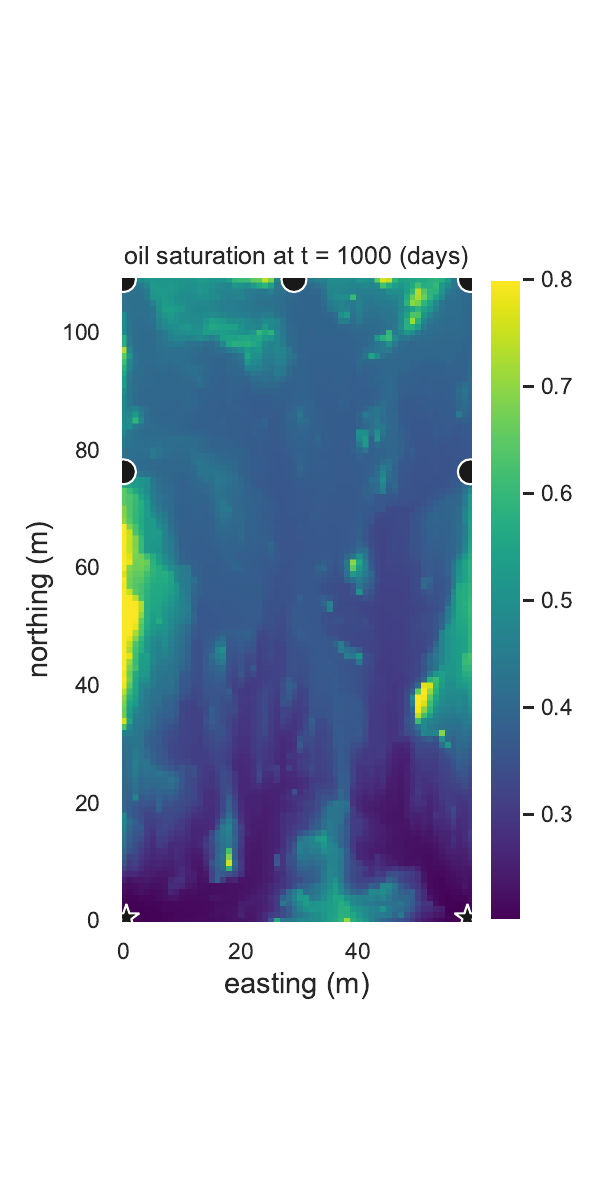}
    \includegraphics[width=\textwidth]{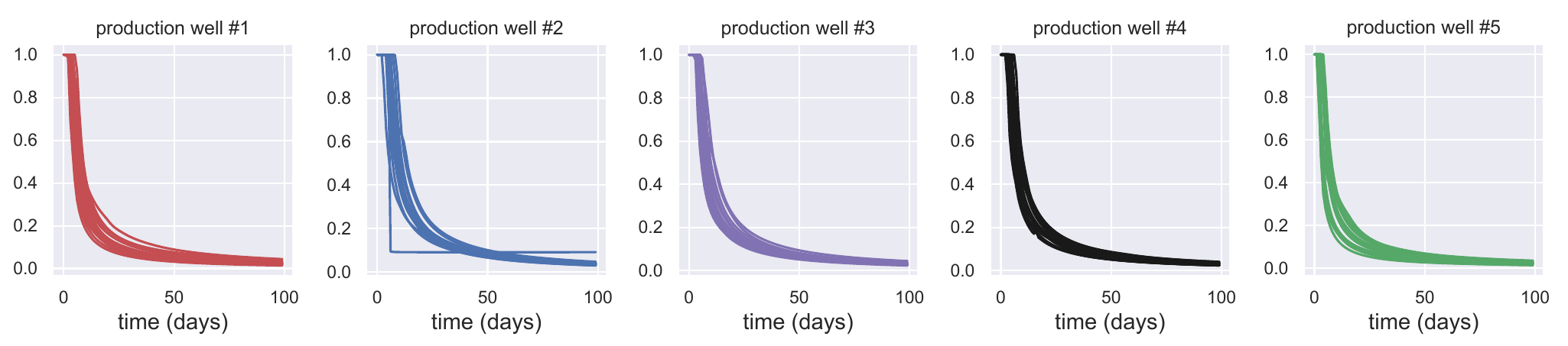}
    \caption{Top: Sample realization of the simulator. The injection wells are depicted with `$\star$' and the production wells are depicted with `$\circ$'. Log permeability is displayed on the left, and oil saturation on the right. Bottom: Fifteen samples of the oil fractional flow rates.}
    \label{fig:sample_run}
\end{figure}

Specifically, we are interested in inferring the unobservable subsurface properties\textemdash the permeability field or the hydraulic conductivities\textemdash of a given 2-D heterogeneous oil reservoir. Its rectangular domain is illustrated in Fig.~\ref{fig:sample_run}. The flow and transport of oil in the reservoir is described by a computational model that simulates immiscible and incompressible two-phase flow (water and oil) in a rectangular domain $B = [L_x, L_y]$ with real dimensions $L_x = 333.58$ and $L_y = 670.56$. For simplicity, gravity effects are ignored. We consider the presence of seven wells, in two of which, water is pumped in, while the oil is pushed out of the remaining five. The two production wells are located at corners of the south boundary of the domain ($(0, 0)$ and $(L_x, 0)$), while the 5 production wells are located on the north size of the of the domain, specifically, at locations $(0, 0.7L_y), (0, L_y), (0.5L_x, L_y), (L_x, L_y)$ and $(L_x, 0.7L_y)$.

We denote with $p_o, p_w, \bv_o, \bv_w$ and $s_o, s_w$ the pressure, velocity and saturation fields of the oil and water respectively where for the later it holds that $s_o + s_w = 1$. Define the total velocity as $\bv = \bv_o + \bv_w$ and the \emph{capillary pressure} as $p_{cap} = p_o - p_w$ which is assumed to be a monotone function of the water saturation $s_w$. The \emph{global pressure} is also defined as $p = p_o - p_c$ where the \emph{complement pressure} $p_c$ is given as 
\begin{equation}
p_c(s_w) = \int_1^{s_w} f_w(s)\frac{\partial p_{cap}}{\partial s_w} ds,
\end{equation}
where $f_w$ is the \emph{fractional-flow} function and measures the water fraction of the total flow and is given by 
\begin{equation}
f_w = \frac{\lambda_w}{\lambda_w + \lambda_o},
\end{equation}
where $\lambda_w$ and $\lambda_o$ are the water and oil total mobilities respectively. These can be expressed as 
\begin{eqnarray}
\lambda_w(s_w) = \frac{\left(s_w^*\right)^2}{\mu_w}, & \lambda_o(s_w) = \frac{\left(1 - s_w^*\right)^2}{\mu_o}, & s_w^* = \frac{s_w - s_{wc}}{1 - s_{or} - s_{wc}},
\end{eqnarray}
where $\mu_o$ and $\mu_w$ are the oil and water viscosities respectively, $s_{or}$ is the \emph{irreducible oil saturation} (lowest oil saturation that can be achieved by displacing oil by water) and $s_{wc}$ is the \emph{connate water saturation} (the saturation of water trapped in the pores of the rock during formation of the rock). Taking the above definitions into consideration, the governing equations of the flow model are~\cite{aarnes_intro}
\begin{subequations}
\begin{align}
\label{eq:elliptic}
-\nabla \cdot \left( \bK \lambda(s_w) \nabla p \right) & = q\\
\label{eq:saturation}
\phi \frac{\partial s_w}{\partial t} + \nabla \cdot\left(f_w(s_w) \bv \right) & = \tilde{q}_w,
\end{align}
\end{subequations}
in the interior of the domain $B$, where $\bK$ is the permeability tensor, $\phi$ is the porosity, $q$ is the source term modeling the water injection and $\tilde{q}_w$ is the source term for the saturation equation given by 
\begin{equation}
\tilde{q}_w = \max\left\{q, 0\right\} + f(s_w)\cdot\min\left\{q, 0\right\}.
\end{equation}
The total velocity is connected to the pressure via the modified Darcy law~\cite{dake}
\begin{equation}
\bv = - \bK\lambda(s_w) \nabla p.
\end{equation}
No-flux boundary conditions are considered on $\partial B$
\begin{equation}
\bv \cdot \mathbf{n} = 0,
\end{equation}
where $\mathbf{n}$ is the unit vector that is normal to the boundary. Furthermore, the initial water saturation is taken to be zero
\begin{equation}
s_w(\bx, t = 0) = 0
\end{equation}
$\forall \bx \in B$.

For this example, we assume a constant porosity $\phi = 10^{-3}$, the water and oil viscosities are $\mu_w = 3\cdot 10^{-4}$ and $\mu_o = 3\cdot 10^{-3}$ while the saturations are $s_wc = s_{or} = 0.2$. The initial boundary value problem is solved using a finite control volume method where for the pressure we use a two-point flux approximation finite volume scheme and for the saturation equation we use a first order upwind scheme~\cite{aarnes_intro}. The discrete system of differential equation is evolved in time using a first order implicit scheme with adaptive step selection upon a Newton-Raphson solver. At last, the wells are modeled via the source term which is set to an injection rate value $\textrm{IR} = 9.3529$~\cite{aarnes_intro}. Each injection well is assigned an injection rate equal to $\mathrm{IR}/2$ whereas the value assigned to each production wells is equal to $-\mathrm{IR} / 5$.

The uncertain input parameters of the model that we seek to infer in this example is the unknown rock permeability $\bK$ that is present throughout the domain. As prior assumptions, we consider $\bK$ to be isotropic, that is $\bK = K \bI$, and $K:= K(\bx, \bxi)$ is modeled as the exponential of a Gaussian random field $G(\bx, \bxi)$ which is expressed by its Karhunen-Lo\`{e}ve expansion,
\begin{equation}
G(\bx, \bxi) = \hat{G}_0 + \gamma\sigma\sum_{i=1}^L\sqrt{\lambda_i}\xi_i \phi_i(\bx),\qquad K(\bx,\bxi) = \exp\left\{G(\bx, \bxi)\right\}\ ,\label{KL:eqn}
\end{equation}
where $\{ \xi_i\}$ are independent Gaussian variables. We make use of the SPE-10 data set~\cite{christie} which is measured in the range of $1200\times 2200\times 170$ ($\textrm{ft}^3$) and is discretized on a regular Cartesian grid with $120\times 220 \times 85$ nodes. We use the $85$ log-fields of dimension $120\times220$ as different realizations from which we compute the sample mean $\hat{G}_0$ and the sample covariance matrix $C(\bx, \bx')$. From the latter we next obtain the eigenvalues and eigenvectors used in equation (\ref{KL:eqn}) and 
we choose $L = 20$ to retain the $20$ most significant terms corresponding to the largest eigenvalues. Note that the relatively small number of samples used to construct the covariance matrix results in a large variance. In order to avoid numerical instabilities in our simulations we have multiplied $\sigma$ by an additional constant taken to be $\gamma = 0.3$ and added an additional constant $\kappa_0 = 10^{-12}$ to $K(\bx, \bxi) = \exp\left\{G(\bx, \bxi)\right\} + \kappa_0$ in order to ensure well posedness of the elliptic problem in (\ref{eq:elliptic}). 

The observable quantities are the oil saturation across the whole domain as well as the oil fractional flow rates at the five production wells
\begin{equation}
    f_{o, i} = \frac{\lambda_o}{\lambda_w + \lambda_o}, i = 1, \dots, 5
\end{equation}
Fig.~\ref{fig:sample_run} shows a sample realization of the oil saturation $s_o$ at time $T$ and 15 samples of the oil fractional-flow rate.

For simplicity, hereafter we denote the log permeability field $G(\bx, \bxi)$ by $\mathbf{G}$, the saturation field $s_o(\bx, t=T)$ by $\mathbf{S}$. Moreover, we denote the flow rate curve (or \emph{oil-cut} curve) of each production well $i$ by $\mathbf{f}_i$, and we use $\mathbf{f}_{i:j}$ to refer to the combination (or Cartesian product) of oil-cut curves of a consecutive series of wells starting from index $i$ and ending at index $j$. Sometimes $\mathbf{f}_i$ may also be expressed as $\mathbf{f}_{i:i}$ so as to unify the notation, depending on the context.

\subsubsection{Outline of Experiment Tasks and Motivations}
\label{sec:outline_exp}
This experiment involves more tasks than the previous one in \S\ref{sec:ko_ode}, in order to more thoroughly explore and reveal the potential of the latent CSGANs in tackling high dimensional inverse problems. For clarity, here we list these tasks out, and highlight their motivations and objectives:
\begin{enumerate}[label=T\arabic*.,ref=\arabic*]
    \item We apply LVAEs to $\mathbf{G}$, $\mathbf{S}$ and $\mathbf{f}_1$, and retrieve the posteriors $p(\mathbf{G}\mid \mathbf{S})$ and $p(\mathbf{G}\mid \mathbf{f}_1)$ with latent CSGANs, then assess their accuracy. This aims to investigate latent CSGANs' ability to handle high dimensional input and output of different formats and unify them under the same latent framework.

    \item We additionally train LVAEs for $\mathbf{f}_{1:2}$, $\mathbf{f}_{1:3}$, $\mathbf{f}_{1:4}$ and $\mathbf{f}_{1:5}$ respectively, and use them for different latent CSGANs to approximate the posteriors $p(\mathbf{G}\mid \mathbf{f}_{1:2})$, $p(\mathbf{G}\mid \mathbf{f}_{1:3})$, $p(\mathbf{G}\mid \mathbf{f}_{1:4})$, $p(\mathbf{G}\mid \mathbf{f}_{1:5})$. Then, along with $p(\mathbf{G}\mid \mathbf{S})$ and $p(\mathbf{G}\mid \mathbf{f}_1)$, we use posterior samples to evaluate their forward propagation error and examine whether there is a \emph{correlation between the posterior's scatter over the support of the distribution and the input condition's intrinsic dimensionality}, as the latter has been obtained by the LVAE's latent space dimension. 
    \label{item:ff}
\end{enumerate}

The probing of this potential correlation in Task~\ref{item:ff} is motivated by the intuitive \emph{preimage theorem}~\cite{guillemin2010differential} (or \emph{regular level set theorem}~\cite{lee2012smooth}):
    \begin{displayquote}
    For \emph{smooth} manifolds $X$ and $Y$, if $y$ is a \emph{regular} value of the \emph{smooth} function $f: X\to Y$, then the preimage $f^{-1}(y)$ is a submanifold of $X$, with $\dim f^{-1}(y) = \dim X - \dim Y$.
    \end{displayquote}
    To explain further the implication of the above theorem, if $f(X)$ is of non-zero measure in $Y$, then \emph{Sard's theorem}~\cite{guillemin2010differential, lee2012smooth} further guarantees that almost every $y\in f(X)\subseteq Y$ is a regular value, which means we may safely apply the preimage theorem to almost every $y$ that can constitute an observed value in practice. Then, for a given deterministic forward function $f$, the approximation of the posterior $p(X\mid Y=y)$ reveals its support, which informs us about the preimage $f^{-1}(y)$\textemdash as $f^{-1}(y)$ supports $p(X\mid Y=y)$. By adopting the manifold hypothesis together with the smoothness assumption on both $X$ and $Y$, the preimage theorem then suggests that the support of $p(X\mid Y=y)$ depends on the dimension of the input condition space $Y$: that is, the higher the dimension of $Y$, the smaller the dimension of $f^{-1}(y)$, which means $p(X\mid Y=y)$ should have smaller variance because $f^{-1}(y)$ is more concentrated in $X$. 
    This correlation, if verified, may carry significant implications for inverse problem solving empowered by Least Volume, as will be discussed later in \S\ref{sec:corr_dim} and \S\ref{sec:conclusion}.
    
\subsubsection{Model Configurations and Training}

To approximate the target posterior distributions of $\mathbf{G}$, we generate a 10000-sample dataset of the triplet $\{(\mathbf{G}, \mathbf{S}, \mathbf{f}_{1:5})\}$ using the aforementioned computational model. Both $\mathbf{G}$ and $\mathbf{S}$ have size $110\times 60$ stemming from the regular Cartesian grid in the 2D space, while each $\mathbf{f}_i$ has 100 dimensions corresponding to 100 equal-paced time steps. To make the data samples more suitable for convolutional neural networks' up- and down-sampling transformations, we use nearest neighbor interpolation to upscale $\mathbf{G}$ and $\mathbf{S}$ to the size $128\times 64$, and $\mathbf{f}_i$ to 128 dimensions. After that, we clip the outlier values in $\mathbf{S}$ (\ie, those not in the range $[0.2, 0.8]$) caused by numerical issues, and normalize all the values of $(\mathbf{G}, \mathbf{S}, \mathbf{f}_{1:5})$ to the range $[0, 1]$. We use 8000 samples as the training set and retain the rest 2000 samples for use as the test set. 

Similar to the KO-ODE experiment, we first use LVAEs to reduce the dimension of $\mathbf{G}$, $\mathbf{S}$ and $\mathbf{f}_{i}$ (or $\mathbf{f}_{i:j}$) respectively, in order to get their compact representation and estimate their dimensionality. For $\mathbf{G}$ and $\mathbf{S}$, we use 2D convolutional LVAEs to autoencode them. Their architectures, hyperparameters and training details are listed in Table~\ref{tab:oil_lvae} and \ref{tab:oil_lvae_hyp}. 
For $\mathbf{f}_{i:j}$ we use 1D convolutional LVAEs for the compression, but the situation is a bit more complicated and requires more effort on fine-tuning the LVAEs' hyperparameters. Specifically, as mentioned in~\cite{qiuyi2024compressing}, the weight $\lambda$ in the objective function (\ref{eq:obj}) controls the trade-off between reconstruction and dimension reduction. If $\lambda$ is too large, then the dimension of the latent space can be reduced arbitrarily by discarding important information (or dimension) of the data, thus increasing the reconstruction error. This is similar to PCA, for which we may reduce the number of principal components to extract, but at the cost of losing precision in data reconstruction. 
For Task~\ref{item:ff} described in \S\ref{sec:outline_exp}, we are interested in retrieving the dimensionalities of different curve combinations $\mathbf{f}_{i:j}$ by taking the LVAE's latent dimensions as their estimates. In order to make the estimates accurate, we need to make the reconstruction errors of different curve combinations comparable with each other. The 1-D convolutional LVAE architectures and hyperparameters in Table~\ref{tab:oil_lvae} and \ref{tab:oil_lvae_hyp} are derived with this consideration taken into account. Their validity in unifying the reconstruction errors will be verified later in \S\ref{sec:oil_lvdr}, Fig.~\ref{fig:oil_rec_f}. 

After training the LVAEs for $\mathbf{G}$, $\mathbf{S}$ and $\mathbf{f}_{i:j}$, we extract their latent spaces and train the latent CSGANs to approximate $p(\mathbf{G}\mid \mathbf{S})$ and $p(\mathbf{G}\mid \mathbf{f}_{i:j})$. Likewise, the training details and architectures are described in Table~\ref{tab:oil_csgan} and \ref{tab:oil_csgan_hyp}. 

\subsubsection{Results: Least Volume Dimension Reduction}
\label{sec:oil_lvdr}
\begin{figure}[bt!]
  \begin{minipage}{.3\textwidth}
    \begin{subfigure}{\textwidth}
      \includegraphics[width=\textwidth]{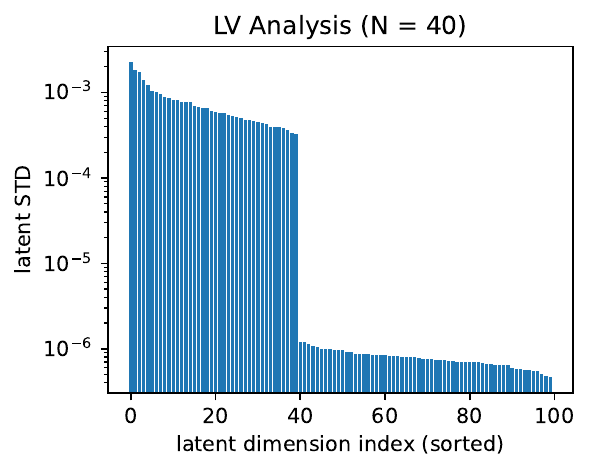}
      \caption{LVAE}
      \label{fig:oil_lv_G}
    \end{subfigure}
    \begin{subfigure}{\textwidth}
      \includegraphics[width=\textwidth]{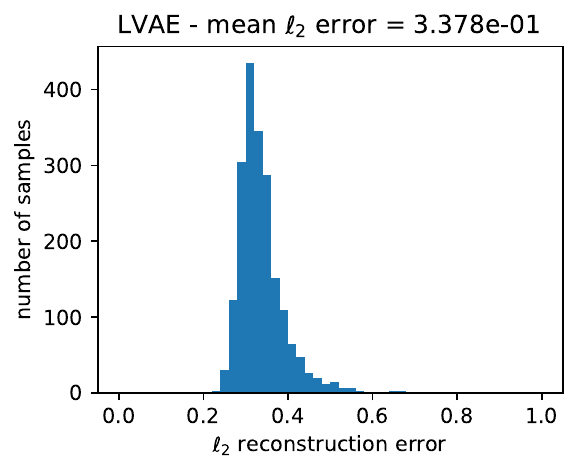}
      \caption{LVAE's $\ell_2$ Error}
      \label{fig:oil_lv_error_G}
    \end{subfigure}
  \end{minipage}
  \hfill
  \begin{minipage}{.3\textwidth}
    \begin{subfigure}{\textwidth}
      \includegraphics[width=\textwidth]{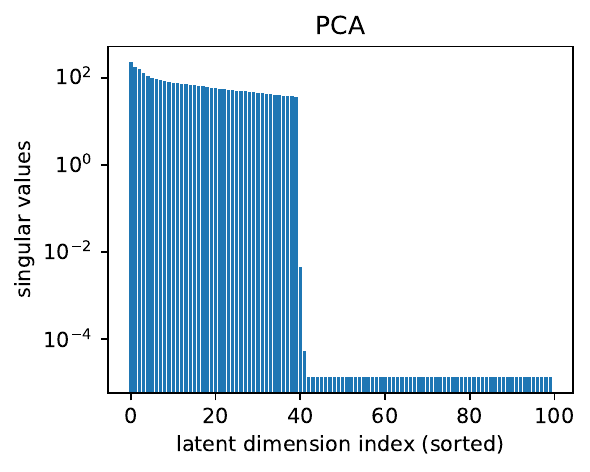}
      \caption{PCA}
      \label{fig:oil_pca_G}
    \end{subfigure}
    \begin{subfigure}{\textwidth}
      \includegraphics[width=\textwidth]{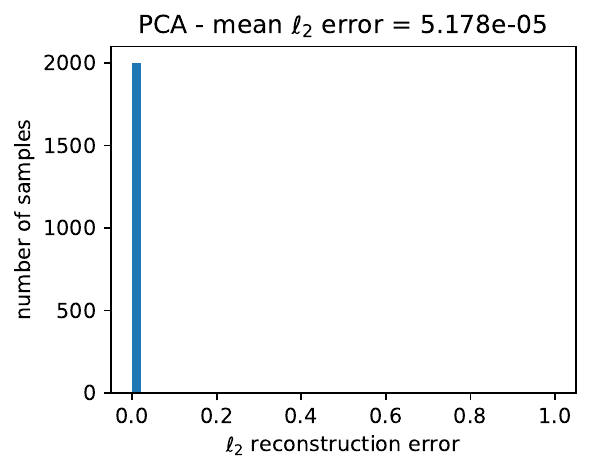}
      \caption{PCA's $\ell_2$ Error}
      \label{fig:oil_pca_error_G}
    \end{subfigure}
  \end{minipage}
  \hfill
  \begin{minipage}{.32\textwidth}
  \vspace{0.2cm}
    \begin{subfigure}{\textwidth}
      \includegraphics[width=\textwidth]{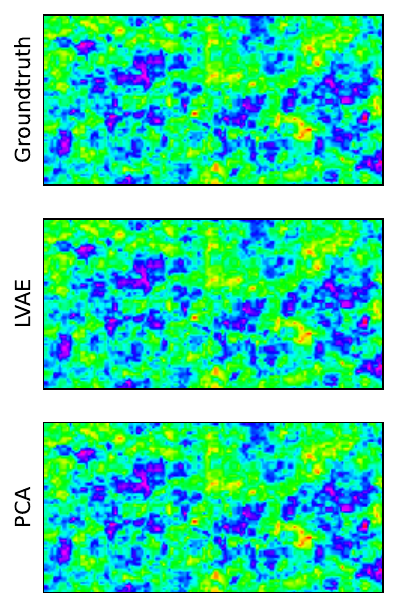}
      \caption{Example of reconstructing $\mathbf{G}$ from 40 latent dimensions. 
      }
      \label{fig:ko_rec}
    \end{subfigure}
  \end{minipage}
  \caption{Dimension reduction results of log permeability field $\mathbf{G}$. By construction, the $\mathbf{G}$ samples are drawn from a Karhunen-Lo\`{e}ve expansion, therefore the linearity of the decomposition allows PCA to achieve reconstruction performance superior to the nonlinear LVAE. However, LVAE also uncovers the 40-D manifold structure of this dataset and has visually indistinguishable reconstruction quality.}
  \label{fig:oil_dr_G}
\end{figure}

\begin{figure}[bt!]
  \begin{minipage}{.3\textwidth}
    \begin{subfigure}{\textwidth}
      \includegraphics[width=\textwidth]{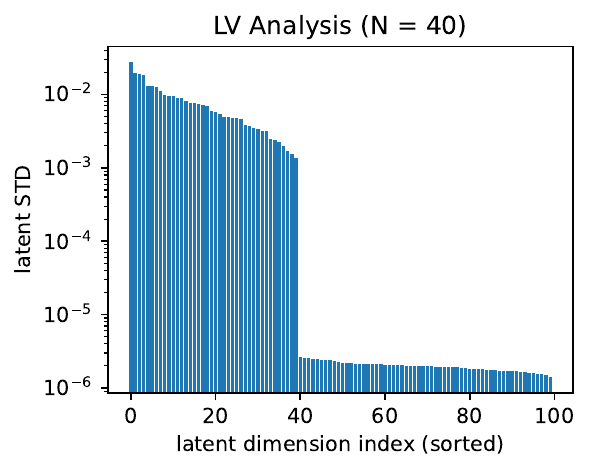}
      \caption{LVAE}
      \label{fig:oil_lv_S}
    \end{subfigure}
    \begin{subfigure}{\textwidth}
      \includegraphics[width=\textwidth]{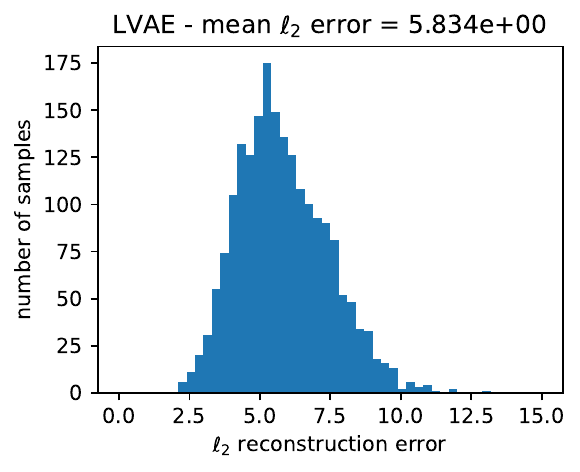}
      \caption{LVAE's $\ell_2$ Error}
      \label{fig:oil_lv_error_S}
    \end{subfigure}
  \end{minipage}
  \hfill
  \begin{minipage}{.3\textwidth}
    \begin{subfigure}{\textwidth}
      \includegraphics[width=\textwidth]{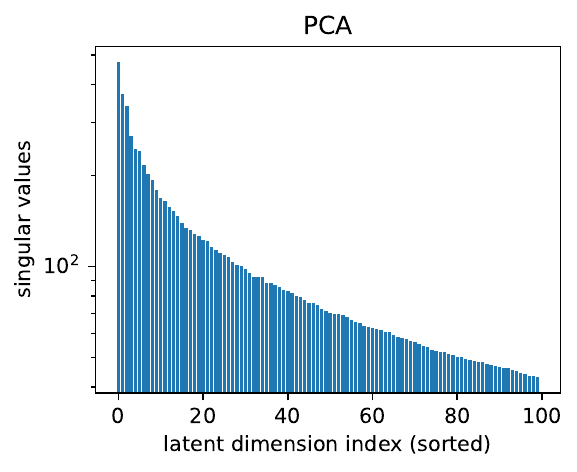}
      \caption{PCA}
      \label{fig:oil_pca_S}
    \end{subfigure}
    \begin{subfigure}{\textwidth}
      \includegraphics[width=\textwidth]{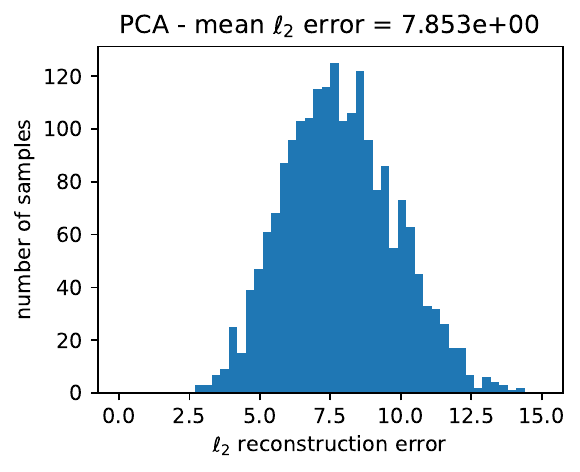}
      \caption{PCA's $\ell_2$ Error}
      \label{fig:oil_pca_error_S}
    \end{subfigure}
  \end{minipage}
  \hfill
  \begin{minipage}{.32\textwidth}
  \vspace{0.2cm}
    \begin{subfigure}{\textwidth}
      \includegraphics[width=\textwidth]{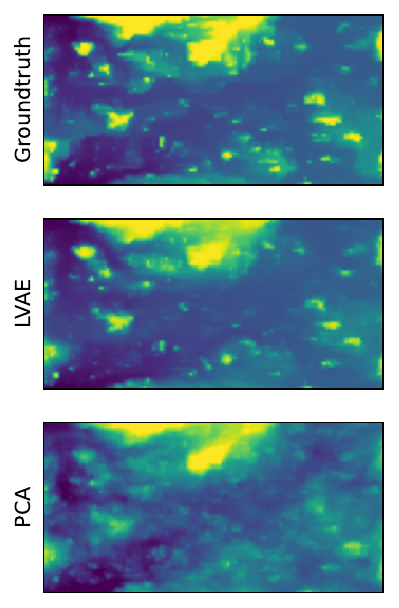}
      \caption{Example of reconstructing $\mathbf{S}$ from 40 latent dimensions. 
      }
      \label{fig:oil_rec_S}
    \end{subfigure}
  \end{minipage}
  \caption{Dimension reduction results of saturation field $\mathbf{S}$. The LVAE can better handle this highly nonlinear dataset than PCA in terms of both compression and reconstruction.}
  \label{fig:oil_dr_S}
\end{figure}

Figure~\ref{fig:oil_dr_G} and \ref{fig:oil_dr_S} demonstrate the dimension reduction results of the LVAEs on the two image datasets $\mathbf{G}$ and $\mathbf{S}$. We can see that the LVAE of the log permeability field $\mathbf{G}$ manages to compress $\mathbf{G}$ into a 40 dimensional latent space, which agrees with $\mathbf{G}$'s groundtruth dimension 40. Because of the inherent linearity of this dataset\textemdash as specified by the Karhunen-Lo\`{e}ve expansion in Equation~\ref{KL:eqn}, both the PCA and the LVAE can achieve comparable reconstruction quality. 

Even more intriguing is that the final saturation field $\mathbf{S}$\textemdash the deterministic governing PDE's final state that depends on $\mathbf{G}$\textemdash is also compressed by its LVAE into a 40 dimensional latent space, and the PCA's poor reconstruction quality shows that the dataset of $\mathbf{S}$ is nonlinear (Fig.~\ref{fig:oil_dr_S}). If this number 40 happens to be $\mathbf{S}$'s actual intrinsic dimension, and we further assume that the causal relationship between $\mathbf{G}$ and $\mathbf{S}$ can be modelled by a smooth map $f_s$ that gives $\mathbf{S} = f_s(\mathbf{G})$, then the preimage theorem tells us the level set $f_s^{-1}(\mathbf{S})$ of almost every $\mathbf{S}$ is a zero-dimensional manifold, \ie, a countable discrete space~\cite{lee2010introduction}. 
Putting this intuitively, it means almost every level set $f_s^{-1}(\mathbf{S})$ comprises only a set of isolated points that do not form lines or surfaces in the space of $\mathbf{G}$, such that the probability density function of $p(\mathbf{G}\mid \mathbf{S})$ can be expressed as the sum of Dirac delta functions. If in addition $f_s$ is injective, then each level set contains only a single point and $p(\mathbf{G}\mid \mathbf{S})$ may even be sufficiently fit by a unimodal regression model with enough complexity. We shall validate this in the next section.

\begin{figure}[bt!]
  \begin{minipage}{.33\textwidth}
    \begin{subfigure}{\textwidth}
      \includegraphics[width=\textwidth]{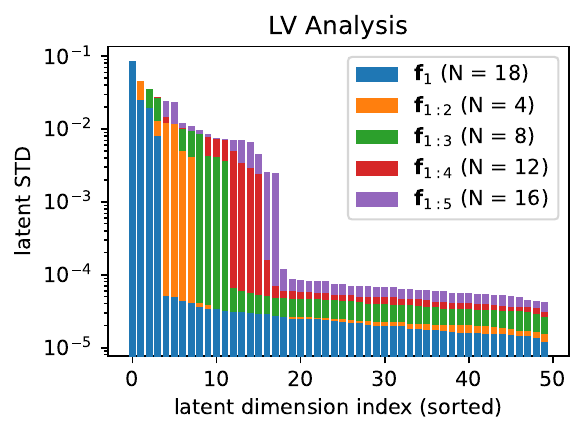}
      \caption{LVAE}
      \label{fig:oil_lv_f}
    \end{subfigure}
    \begin{subfigure}{\textwidth}
      \includegraphics[width=\textwidth]{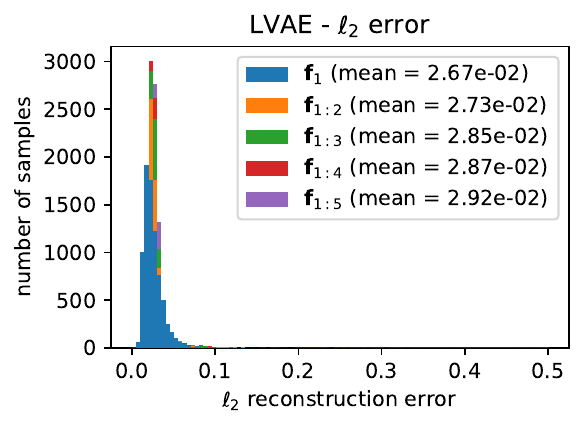}
      \caption{LVAE's $\ell_2$ Error}
      \label{fig:oil_lv_error_f}
    \end{subfigure}
  \end{minipage}
  \hfill
  \begin{minipage}{.33\textwidth}
    \begin{subfigure}{\textwidth}
      \includegraphics[width=\textwidth]{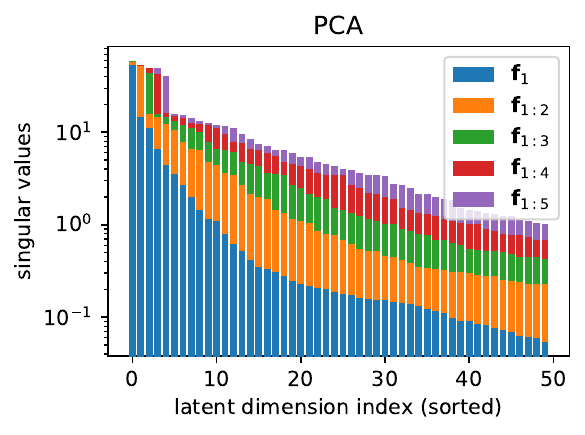}
      \caption{PCA}
      \label{fig:oil_pca_f}
    \end{subfigure}
    \begin{subfigure}{\textwidth}
      \includegraphics[width=\textwidth]{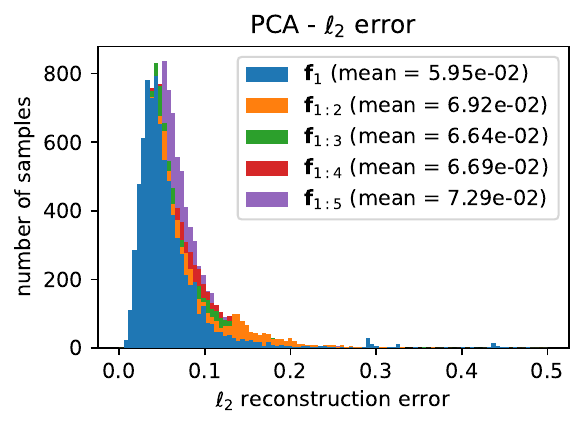}
      \caption{PCA's $\ell_2$ Error}
      \label{fig:oil_pca_error_f}
    \end{subfigure}
  \end{minipage}
  \hfill
  \begin{minipage}{.32\textwidth}
  \vspace{0.4cm}
    \begin{subfigure}{\textwidth}
      \includegraphics[width=\textwidth]{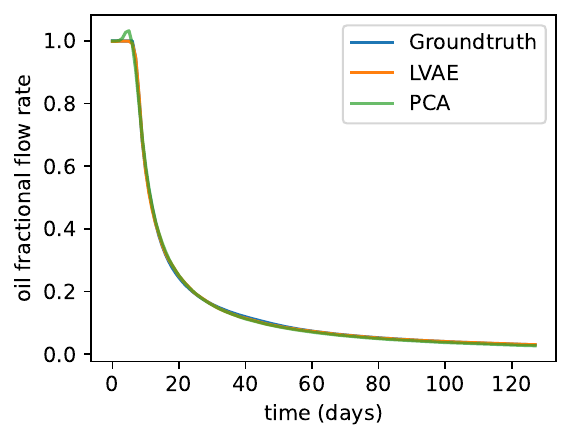}
      \vfill
      \includegraphics[width=\textwidth]{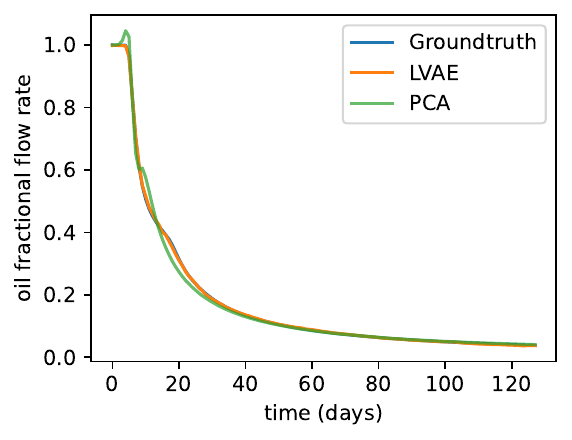}
      \caption{Examples of reconstructing $\mathbf{f}_i$
      }
      \label{fig:oil_rec_f}
    \end{subfigure}
  \end{minipage}
  \caption{Dimension reduction results of oil-cut curves $\mathbf{f}_{1:j}$. The LVAEs are tuned to have comparable $\ell_2$ reconstruction errors for each production well. When using the same number of latent dimensions, PCA's reconstruction is not as good as that of the LVAEs, but is still acceptable in most cases. This means the $\mathbf{f}_{i:j}$ datasets are not highly nonlinear.} 
  \label{fig:oil_dr_f}
\end{figure}

The LVAEs also obtain some interesting results on the oil-cut curve combinations $\mathbf{f}_{1:j}$ for $j\in \{1,2,3,4,5\}$. As we can see in Fig.~\ref{fig:oil_dr_f}, starting from $\mathbf{f}_{1:1}$ to $\mathbf{f}_{1:4}$, every time we include an additional oil-cut curve in $\mathbf{f}_{1:j}$, the curve combination $\mathbf{f}_{1:j}$'s dimension increases by 4. This not only suggests that each oil-cut curve $\mathbf{f}_{i}$ lives on a 4-D manifold\textemdash which can be double-checked by applying LVAEs to $\mathbf{f}_j$ other than $\mathbf{f}_1$\textemdash but also implies that $\mathbf{f}_i$ for $i\in\{1,2,3,4\}$ are uncorrelated with each other, as otherwise we should be able to use less dimensions to represent some of these combinations. 
However, the situation differs as $j$ increases to 5. The result shows that $\mathbf{f}_{1:5}$ contains only 2 extra dimensions compared to $\mathbf{f}_{1:4}$, while $\mathbf{f}_5$ alone has 4 dimensions, indicating that $\mathbf{f}_5$ correlates with the other oil-cut curves. This hints that the amount of information provided by the production wells starts to ``saturate''. The same ``saturation effect" is also perceived on $\mathbf{S}$: each pixel of $\mathbf{S}$ has its own variation on $\mathbb{R}$ and thus has 1 dimension, but the Cartesian product of these $128\times 64$ pixels only has 40 dimensions rather than 8192 due to the correlation among them. 

This information (or dimension) saturation phenomenon is insightful. As an example, in the scenario where we are provided with an unlimited number of sensors, if we want to use them to monitor the status of a structure, it may not be true that the more sensors we mount, the more thoroughly we can grasp the structure's condition, given that there probably exists a threshold on the number of sensors to mount on the structure, beyond which the sensor data's dimension cannot increase to provide more information about the structure, while the sensors' cost becomes staggering. Monitoring the sensor data's intrinsic dimension with LVAE may help us determine the threshold.

\subsubsection{Results: Bayesian Inference of Log Permeability Field via $p(\mathbf{G}\mid\mathbf{S})$}
\label{sec:oil_cgan_S}
With the LVAEs at hand, we then train the latent CSGANs to approximate the posteriors $p(\mathbf{G}\mid\mathbf{S})$ and $p(\mathbf{G}\mid\mathbf{f}_{1:j})$ for $j\in\{1,2,3,4,5\}$. The two key aspects for assessing the quality of the approximations, in accordance with \S\ref{sec:outline_exp}, are:
\begin{itemize}
    \item The \emph{accuracy} of their approximation, \ie, how well does their conditional pushforward measure $\mathbb{P}_g$ align with the groundtruth posterior probability measure $\mathbb{P}_r$. We will investigate their accuracy in both the output $\mathbf{G}$ space and the input condition spaces using different metrics to obtain a comprehensive understanding, as some metrics may not work under certain circumstances due to the complexity of this problem. 
    
    \item The \emph{uncertainty} around different posteriors, and the uncertainty's correlation with the input condition's intrinsic dimension. In \S\ref{sec:corr_dim}, we shall use the computational entropy estimator powered by the KSG method~\cite{kraskov2004estimating} to quantify the uncertainty of different posterior distributions over the output space of $\mathbf{G}$. The larger the entropy, the higher the uncertainty. 
\end{itemize}

\begin{figure}[bt!]
     \centering
     \begin{subfigure}[b]{\textwidth}
         \centering
         \includegraphics[width=\textwidth, trim={0 3.5cm 0 0.7cm},clip]{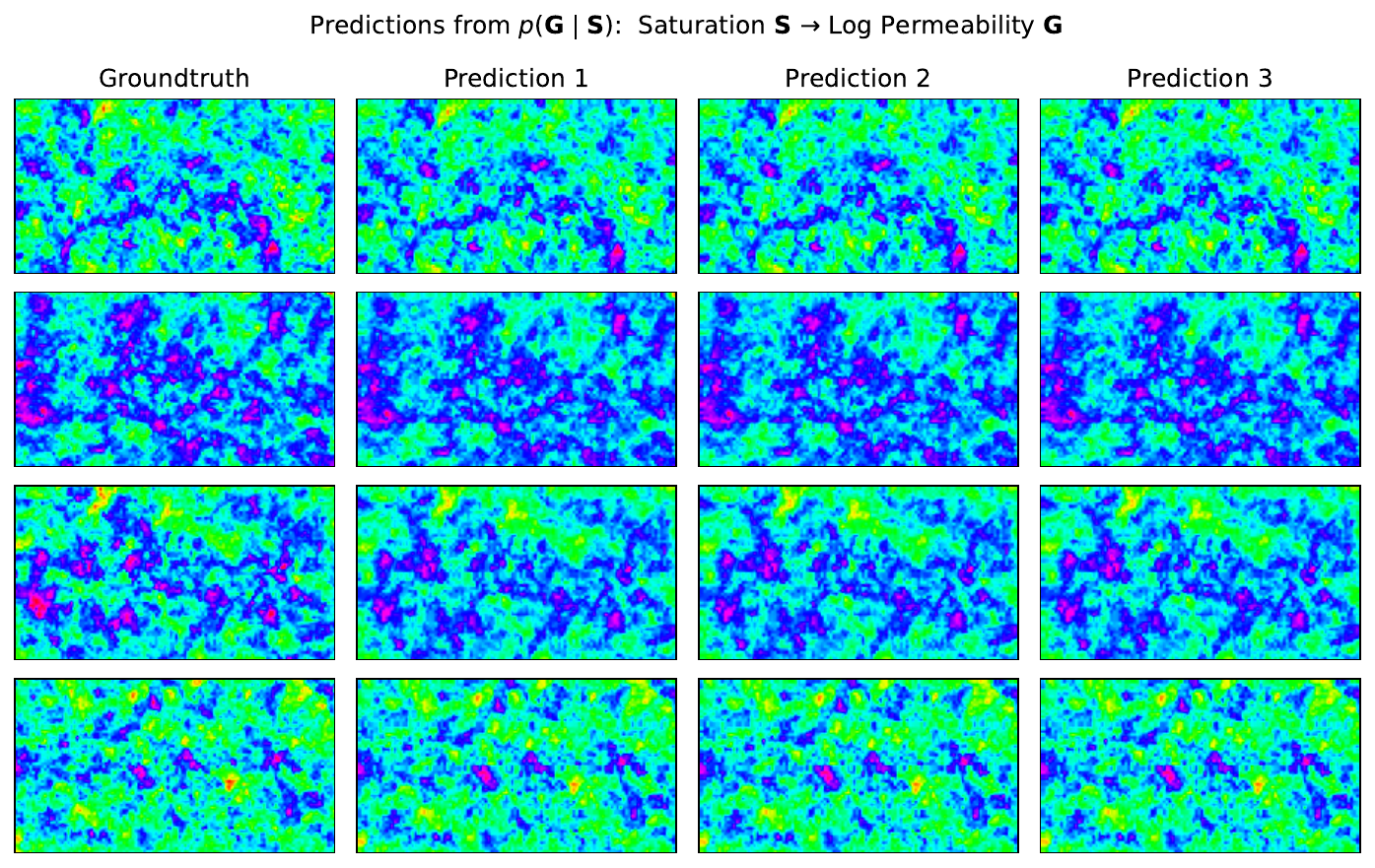}
         \caption{$\mathbf{G}$ space: Predictions from latent CSGAN for $p(\mathbf{G}\mid\mathbf{S})$}
         \label{fig:oil_cgan_pred_S}
     \end{subfigure}
     \\
     \begin{subfigure}[b]{\textwidth}
         \centering
         \includegraphics[width=\textwidth, trim={0 0 0 0.7cm},clip]{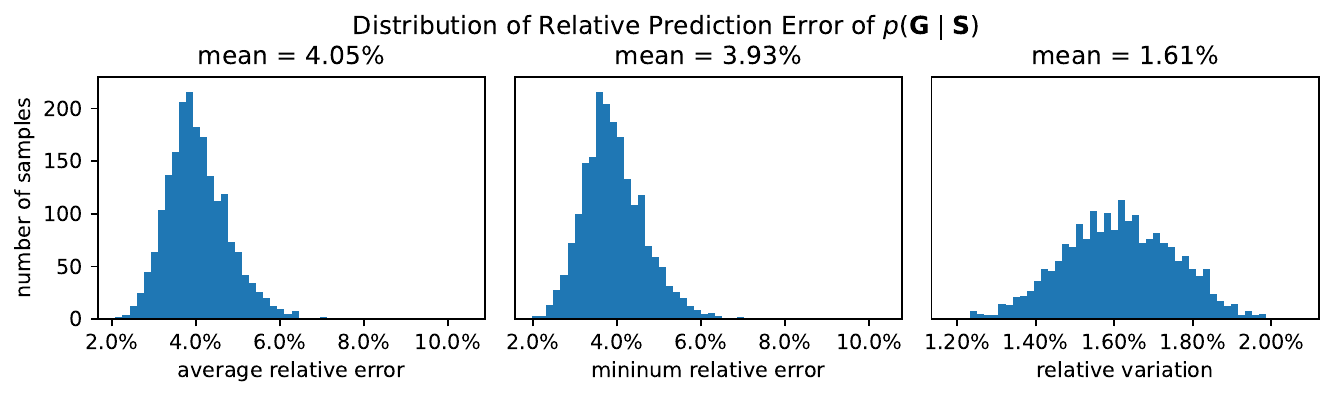}
         \caption{$\mathbf{G}$ space: Error and variation distribution of latent CSGAN for $p(\mathbf{G}\mid\mathbf{S})$}
         \label{fig:oil_cgan_error_S}
     \end{subfigure}
     \\
     \begin{subfigure}[b]{0.7\textwidth}
         \centering
         \includegraphics[width=\textwidth, trim={0 3.5cm 5.8cm 0.7cm},clip]{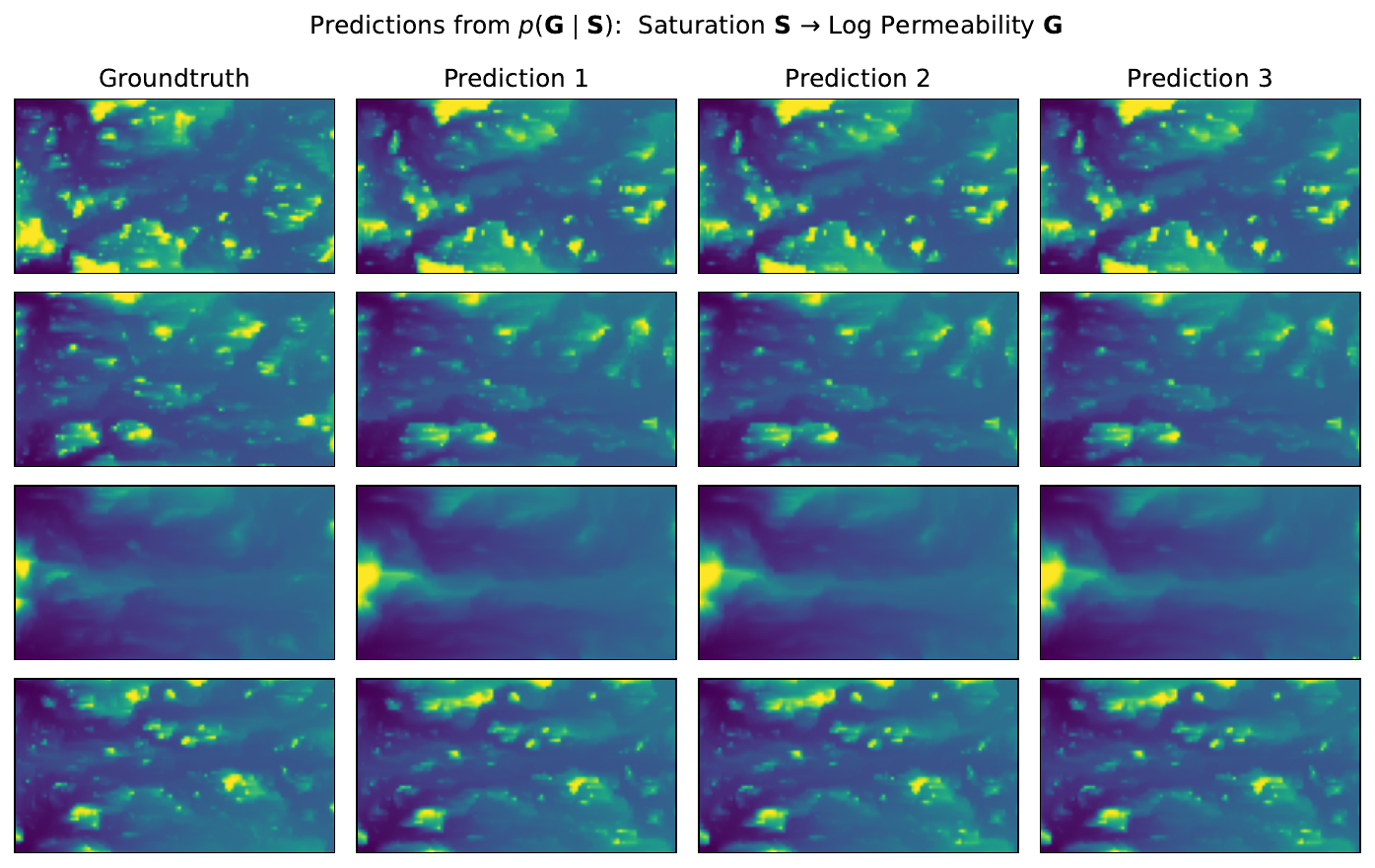}
         \caption{$\mathbf{S}$ space: Saturation field $\hat{\mathbf{S}}$ of predicted $\hat{\mathbf{G}}$ against groundtruth $\mathbf{S}$}
         \label{fig:oil_cgan_pred_s_S}
     \end{subfigure}
     \hfill
     \begin{subfigure}[b]{0.29\textwidth}
         \centering
         \includegraphics[width=\textwidth, trim={0 0.3cm 0 0},clip]{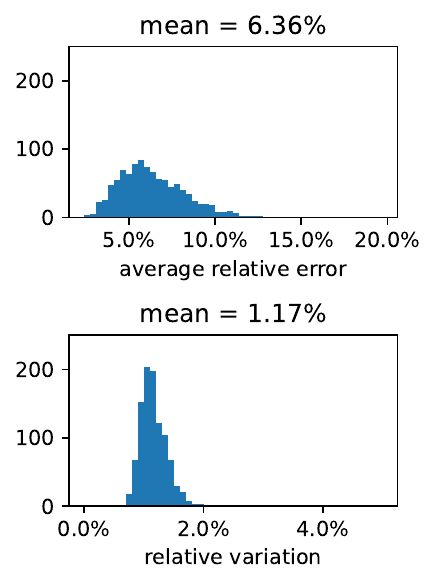}
         \caption{$\mathbf{S}$ space: Error and variation distribution of $\hat{\mathbf{S}}$}
         \label{fig:oil_cgan_error_s_S}
     \end{subfigure}
     \caption{Predictions and errors of latent CSGAN for $p(\mathbf{G}\mid\mathbf{S})$ in $\mathbf{G}$ and $\mathbf{S}$ space}
     \label{fig:oil_cgan_S}
\end{figure}

First, we inspect the latent CSGAN's learning accuracy of $p(\mathbf{G}\mid \mathbf{S})$ in this section. Figure~\ref{fig:oil_cgan_S} studies and illustrates it from different perspectives. Figure~\ref{fig:oil_cgan_pred_S} displays the predicted $\mathbf{G}$ made by the CSGAN on three randomly chosen test cases, juxtaposed against their groundtruth $\mathbf{G}$ on the left. For each case, given the input condition $\mathbf{S}$, the generator stochastically produces three $\mathbf{G}$ predictions for illustration. We can see that the predictions are almost visually indistinguishable from their groundtruths, and in each test case, the predictions also look identical to each other. This means the CSGAN is accurate, and its target posterior $p(\mathbf{G}\mid \mathbf{S})$ has very low uncertainty, which agrees well with our projection in the beginning of \S\ref{sec:oil_lvdr}. 

The results in Fig.~\ref{fig:oil_cgan_error_S} bolster the above claim quantitatively. There are two metrics involved: \emph{relative error} and \emph{relative variation}. They are respectively \emph{length-scale-independent} quantities for evaluating each prediction's deviation from its groundtruth, and the variation of a batch of predictions from each other. They are defined as follows:
\begin{itemize}
    \item \textit{Length Scale:} The log permeability's \emph{length scale} $c_G = 6\sigma$ is evaluated by taking the standard deviation $\sigma$ among all the pixels of all the 2000 $\mathbf{G}$ samples, then scaling it up by 6. This $6\sigma$ evaluation makes sense because the log permeability values are generated from a Gaussian random field, and the $6\sigma$ range encloses around $99.7\%$ log permeability values while ignoring the few outliers on the long tail, avoiding underestimating the quantities below.
    
    \item \textit{Relative Error:} For each prediction, the absolute error between $\hat{\mathbf{G}}$ and $\mathbf{G}$ are first evaluated pixel-wisely, which gives a $110\times 60$ grid of errors denoted by $\mathbf{E} = |\hat{\mathbf{G}} - \mathbf{G}|$. Then the mean absolute error $\bar{e}$ is evaluated over all pixels of $\mathbf{E}$.  Finally the \emph{relative error} $\epsilon$ of each prediction $\hat{\mathbf{G}}$ is defined by $\epsilon = \bar{e}/c_G$. We choose this pixel-wise $\ell_1$-distance-based error over the $\ell_2$ distance (as in \S\ref{sec:ko_cgan}, Fig.~\ref{fig:ko_cgan_error_all}) to assess the deviation between image data because of not only its intuitiveness, but also the $\mathbf{G}$ space's high dimensionality, as the $\ell_p$ distance with larger $p$ in general works worse and less intuitively in higher dimensional spaces due to the diminishing contrast~\cite{beyer1999nearest, aggarwal2001surprising}. Since $\epsilon$ is essentially a ratio, we express it in Fig.~\ref{fig:oil_cgan_S} in percentage.

    \textit{Average and Minimum Relative Error:}
    For each $\mathbf{G}$ and $\mathbf{S}$ of the 2000 test cases, we sample 50 predictions $\hat{\mathbf{G}}^{(j)}$ for $j\in\{1,\cdots,50\}$ using the CSGAN. Then we evaluate the average and minimum of $\epsilon$ across these 50 predictions. They provide a sense of the average prediction accuracy in each test case, and the best accuracy CSGAN can achieve out of all the nonidentical predictions. 
    
    \item \textit{Relative Variation:}
    For each test case, we evaluate the value range (\ie, $\mathrm{max}-\mathrm{min}$) of each pixel of $\hat{\mathbf{G}}$ across the 50 predictions, then take the average of these ranges across all pixels and divide it by the length scale $c_G$ to obtain the \emph{relative variation} of predictions in each test case. This value provides us a crude sense of the concentration of these predictions\textemdash or the uncertainty of the pushforward measure $\mathbb{P}_g^{\mathbf{G}\mid \mathbf{S}}$, serving as an appetizer prior to the entropy-based discussion of uncertainty in \S\ref{sec:corr_dim}. Likewise, we express the relative variation in percentage in Fig.~\ref{fig:oil_cgan_S}.
\end{itemize}
Figure~\ref{fig:oil_cgan_error_S} shows that the predictions in each test case have very high accuracy and concentration, which agrees with our visual observation in Fig.~\ref{fig:oil_cgan_pred_S}. The proximity between the average and the minimum relative error further testifies to the low uncertainty of $\mathbb{P}_g^{\mathbf{G}\mid \mathbf{S}}$.

However, so far we have only done inspections in the output space of $\mathbf{G}$. It may happen that the mapping $f_s:\mathbf{G}\mapsto\mathbf{S}$\textemdash as introduced in \S\ref{sec:oil_lvdr} \textemdash has large Lipschitz constant, such that even if a prediction $\hat{\mathbf{G}}$ for condition $\mathbf{S}$ has a small deviation from its groundtruth $\mathbf{G}$, its saturation field $\hat{\mathbf{S}} \coloneqq f_s(\hat{\mathbf{G}})$ still differs largely from the groundtruth input condition $\mathbf{S}$. Fortunately, this is not the case. As Fig.~\ref{fig:oil_cgan_pred_s_S} and \ref{fig:oil_cgan_error_s_S} show, the saturation fields of the predicted log permeability fields\textemdash obtained after simulation\textemdash are very close to their groundtruth saturation fields, and they also have very small variations. It should be noted that for the relative error and relative variation in Fig.~\ref{fig:oil_cgan_error_s_S}, we use the actual range $0.6 = 0.8-0.2$ of the saturation values as the length scale instead of using its standard deviation, as the saturation does not follow a Gaussian distribution from our observation, and the $6\sigma$ evaluation can lead to a larger length scale value overestimating the accuracy. Moreover, due to the expensive cost of the simulation, we only perform this investigation over 1000 test cases, and for each of them we only simulate the saturation field for 5 predictions rather than 50. Nonetheless, we believe this relatively limited evaluation in the $\mathbf{S}$ space can still sufficiently reflect the accuracy of $\hat{\mathbf{S}}$.

This precision in both the $\mathbf{G}$ space and the $\mathbf{S}$ space, together with the minimization of the Sinkhorn divergence during training, shows that the latent CSGAN approximates the target posterior $p(\mathbf{G}\mid \mathbf{S})$ sufficiently, so we may expect the true $p(\mathbf{G}\mid \mathbf{S})$ to have these observed characteristics. If this holds, then as Fig.~\ref{fig:oil_cgan_S} shows, $p(\mathbf{G}\mid \mathbf{S})$ is indeed a distribution with very low uncertainty, and is almost surely unimodal, since for each $\mathbf{S}$ it practically always yields the same $\mathbf{G}$. This confirms our projection in \S\ref{sec:oil_lvdr}, and further reveals that $f_s: \mathbf{G}\mapsto\mathbf{S}$ is not only injective but also very likely \emph{homeomorphic}~\cite{lee2010introduction}, given that the generator of CSGAN is a continuous function that can approximate $f_s^{-1}$ well, which means $f_s^{-1}$ could be continuous.

\subsubsection{Results: Bayesian Inference of Log Permeability Field via $p(\mathbf{G}\mid\mathbf{f}_{1:j})$}
\label{sec:oil_cgan_f}

\begin{figure}[bt!]
     \centering
     \begin{subfigure}[b]{\textwidth}
         \centering
         \includegraphics[width=\textwidth, trim={0 0.2cm 0 0.7cm},clip]{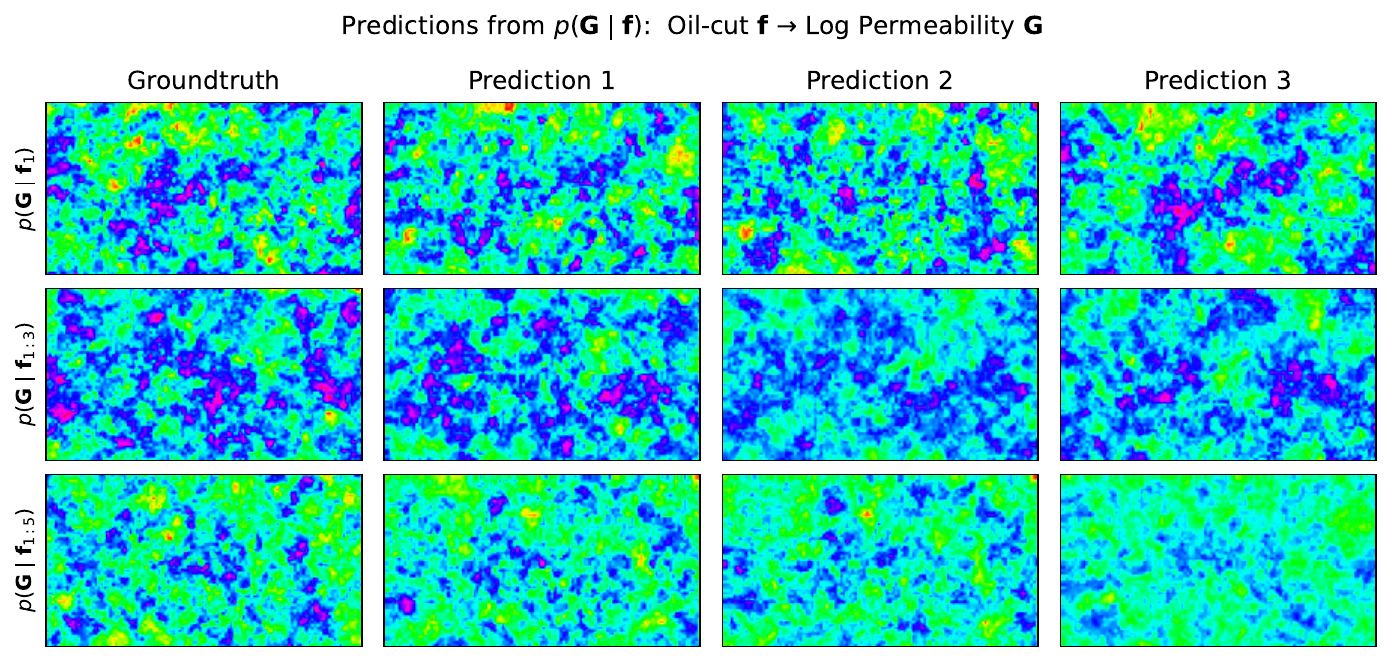}
         \caption{$\mathbf{G}$ space: Predictions from latent CSGAN for $p(\mathbf{G}\mid\mathbf{f}_{1:j})$ with $j\in\{1,3,5\}$}
         \label{fig:oil_cgan_pred_f}
     \end{subfigure}
     \\
     \begin{subfigure}[b]{\textwidth}
         \centering
         \includegraphics[width=\textwidth, trim={0 0 0 0},clip]{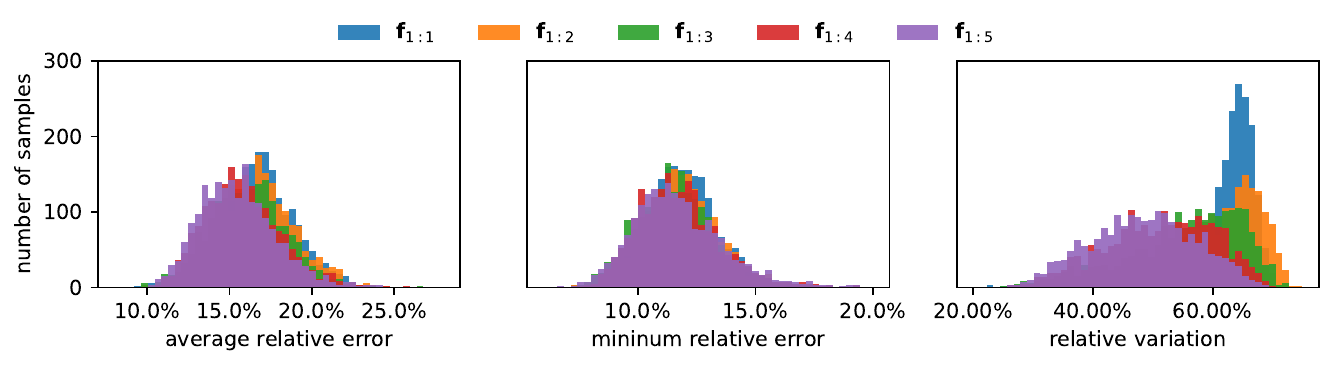}
         \caption{$\mathbf{G}$ space: Error distribution of latent CSGAN for $p(\mathbf{G}\mid\mathbf{f}_{1:j})$ with $j\in\{1,2,3,4,5\}$}
         \label{fig:oil_cgan_error_f}
     \end{subfigure}
     \\
     \begin{subfigure}[b]{0.31\textwidth}
         \centering
         \includegraphics[width=\textwidth, trim={0 0.25cm 0 0},clip]{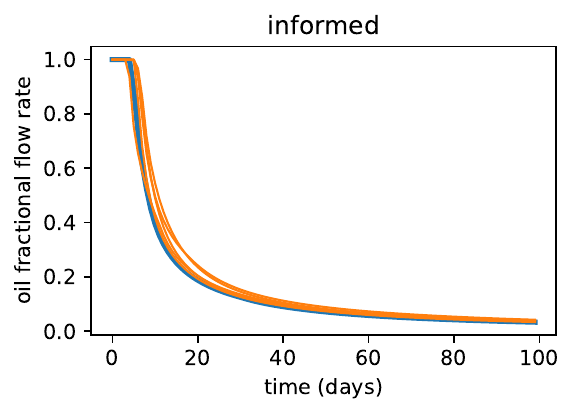}
         \includegraphics[width=\textwidth, trim={0 0.25cm 0 0},clip]{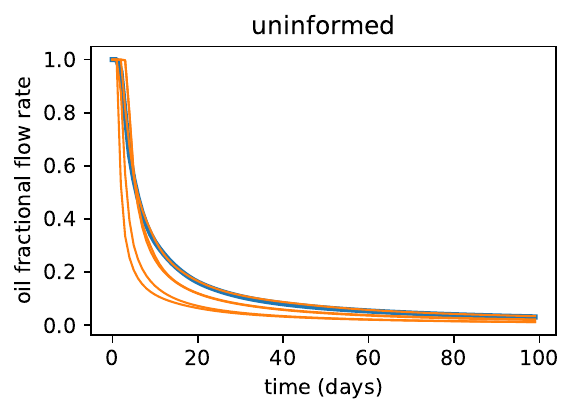}
         \caption{$\mathbf{f}$ space: $\hat{\mathbf{f}}_i$ (orange) vs $\mathbf{f}_i$}
         \label{fig:oil_cgan_pred_f_f}
     \end{subfigure}
     \hfill
     \begin{subfigure}[b]{0.68\textwidth}
         \centering
         \includegraphics[width=\textwidth, trim={0 0 0 0},clip]{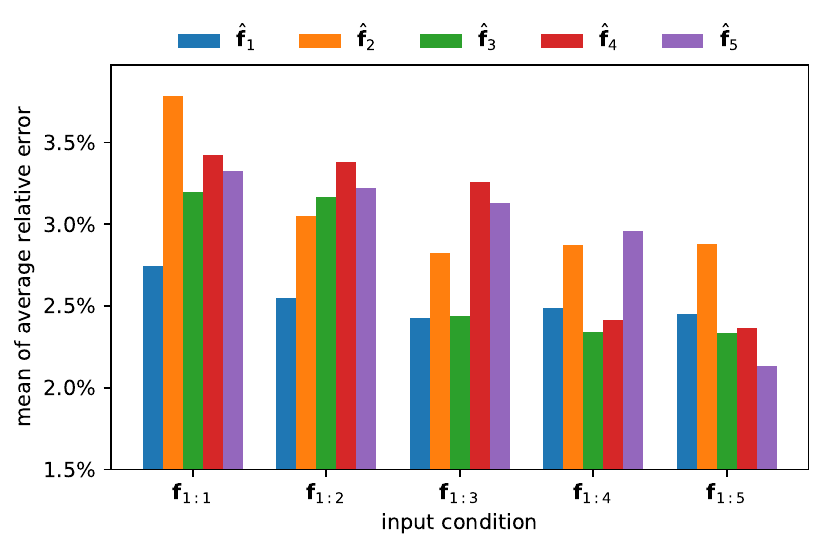}
         \caption{$\mathbf{f}$ space: Saturation field $\hat{\mathbf{S}}$ of predicted $\hat{\mathbf{G}}$ against groundtruth $\mathbf{S}$}
         \label{fig:oil_cgan_error_f_f}
     \end{subfigure}
     \caption{Predictions and errors of latent CSGANs for $p(\mathbf{G}\mid\mathbf{f}_{1:j})$ in $\mathbf{G}$ and $\mathbf{f}$ space}
     \label{fig:oil_cgan_f}
\end{figure}

Next, we perform a series of inspections on the latent CSGANs approximating $p(\mathbf{G}\mid\mathbf{f}_{1:j})$ in a manner similar to the previous section: that is, we inspect the CSGANs' accuracy in both the output space of $\mathbf{G}$ and the input spaces of $\mathbf{f}_{1:j}$. The results are illustrated in Fig.~\ref{fig:oil_cgan_f}.

Figure~\ref{fig:oil_cgan_pred_f} shows the predictions made by the CSGANs' fitting $p(\mathbf{G}\mid \mathbf{f}_{1})$, $p(\mathbf{G}\mid \mathbf{f}_{1:3})$ and $p(\mathbf{G}\mid \mathbf{f}_{1:5})$. Here we omit the CSGANs fitting $p(\mathbf{G}\mid \mathbf{f}_{1:2})$ and $p(\mathbf{G}\mid \mathbf{f}_{1:4})$, for both space reasons and the fact that their predictions have similar visual patterns to the other three. In contrast to Fig.~\ref{fig:oil_cgan_pred_S}, we can see that the predictions in each test case no longer look identical to each other, and they do not match the groundtruth. This can be anticipated, as \S\ref{sec:oil_lvdr} reveals that the oil-cut curves $\mathbf{f}_{1:j}$ do not have as many intrinsic dimensions as that of the saturation field $\mathbf{S}$, so the preimage theorem\textemdash if still applicable here\textemdash implies that $\mathbf{f}_{1:j}$'s preimages in the $\mathbf{G}$ space should be submanifolds of much higher dimensions ($22\sim 36$ from LVAEs' estimates) than the 0-dimensional $f_s^{-1}(\mathbf{S})$. Therefore, in each test case, there seems to be a large scatter around the groundtruth $\mathbf{G}$ of the given $\mathbf{f}_{1:j}$,  for which the log permeability fields can satisfactorily produce the observed $\mathbf{f}_{1:j}$, resulting in an overall high variance of the posterior density.

Because the visual inspection can no longer help grasp the accuracy and uncertainty of CSGANs, we then try the relative error and relative variation to quantitatively probe the CSGANs, as shown in Fig.~\ref{fig:oil_cgan_error_f}. Similar to \S\ref{sec:oil_cgan_S}, for each test case we produce 50 predictions to evaluate these quantities. It appears that all the CSGANs have comparable average and minimum relative errors, and these errors are much larger than their counterparts in Fig.~\ref{fig:oil_cgan_error_S}, which just reaffirms our observation in Fig.~\ref{fig:oil_cgan_pred_f} and concludes nothing more about the CSGANs' accuracy. Also, we should not expect these 50 predictions to greatly drag down the minimum relative error to a level comparable to that in Fig.~\ref{fig:oil_cgan_error_S}. This is because the preimages of $\mathbf{f}_{1:j}$ probably constitutes $22\sim36$-D manifolds, but due to the curse of dimensionality, 50 samples can only cover a negligible part of each, so in each test case, it is not likely that one of the predictions will fall into the groundtruth $\mathbf{G}$'s vicinity. 
However, the distribution of relative variation reveals a pattern anticipated in \S\ref{sec:outline_exp}: the more intrinsic dimensions in the input condition, the larger the variance of the posterior. It shows that the CSGANs taking more $\mathbf{f}_i$ as input have less variation in their predictions. The preimage theorem sheds some light on this phenomenon: the dimensionality of $\mathbf{f}_{1:j}$'s preimage\textemdash which comprises all the valid predictions\textemdash decreases as $j$ increases. 

Investigating the $\mathbf{G}$ space alone thus cannot determine the accuracy of these CSGANs, so we switch to the input condition space of $\mathbf{f}_{1:j}$. In order to assess whether or not the predicted log permeability fields $\hat{\mathbf{G}}$ are in or, practically, near the preimages of $\mathbf{f}_{1:j}$, for each of 1000 selected test cases, we simulate the corresponding oil-cut curves $\hat{\mathbf{f}}_{1:j}^{(k)}$ of 5 predictions $\hat{\mathbf{G}}^{(k)}$ with $k\in\{1,2,3,4,5\}$ and compare them with the groundtruth $\mathbf{f}_{1:j}$ both visually and quantitatively as in \S\ref{sec:oil_cgan_S}. Figure~\ref{fig:oil_cgan_pred_f_f} shows an example case predicted by the CSGAN learning $p(\mathbf{G}\mid\mathbf{f}_{1})$. Here the figure on the top shows $\mathbf{f}_1$ along with the $\hat{\mathbf{f}}_1^{(k)}$ derived from its five predicted $\hat{\mathbf{G}}^{(k)}$, and the one on the bottom shows $\mathbf{f}_5$ along with the  five $\hat{\mathbf{f}}_5^{(k)}$. We can see this CSGAN\textemdash informed only by $\mathbf{f}_1$ and not by $\mathbf{f}_5$\textemdash produces $\hat{\mathbf{G}}^{(k)}$ that results in oil-cut curves that match $\mathbf{f}_1$ better than $\mathbf{f}_5$. Figure~\ref{fig:oil_cgan_error_f_f} further plots the error distributions in the $\mathbf{f}$ spaces to provide a more comprehensive view. The $x$-axis corresponds to the five latent CSGANs taking different number of oil-cut curves $\mathbf{f}_{1:j}$ as input condition, the $y$-axis shows the \emph{mean} of the average relative errors of $\hat{\mathbf{f}}_i$ against $\hat{\mathbf{f}}_i$ evaluated across the 1000 test cases, and different colors in Fig.~\ref{fig:oil_cgan_error_f_f} represent the relative errors at different production wells. We notice a clear pattern that whenever the oil-cut $\mathbf{f}_i$ at a given well $i$ is used to inform certain CSGANs as an input condition, those CSGANs generate predictions $\hat{\mathbf{G}}$ which in general yield $\hat{\mathbf{f}}_i$ that match $\mathbf{f}_i$ better. This indicates that the CSGANs indeed have learned to produce $\hat{\mathbf{G}}$ close to the preimage of its input condition $\mathbf{f}_{1:j}$. However, the learning accuracy remains poorer compared to the $p(\mathbf{G}\mid\mathbf{S})$ case of \S\ref{sec:oil_cgan_S}, Fig.~\ref{fig:oil_cgan_S}. The standard deviations of these well-wise average relative errors in Fig.~\ref{fig:oil_cgan_error_f_f}\textemdash evaluated across the 1000 test cases\textemdash turn out to be all around $1.5\%$, which is considerable compared to the errors' mean values typically around $2.5\%$. This means the CSGANs are not able to consistently generate accurate $\hat{\mathbf{G}}$ for different $\mathbf{f}_{1:j}$ among the test cases.

\subsubsection{Results: Correlation between Uncertainty of Posterior and Intrinsic Dimension of Input Condition}
\label{sec:corr_dim}

The distributions of relative variations in Fig.~\ref{fig:oil_cgan_error_S} and \ref{fig:oil_cgan_error_f} have already given us an idea about the correlation between the input's dimensionality and the output's uncertainty as anticipated in \S\ref{sec:outline_exp}. Here, we further confirm it with the more rigorous metric\textemdash \emph{entropy}~\cite{cover2006info}.
The differential entropy of the latent CSGANs' pushforward measure cannot be evaluated analytically, since the density function's closed-form expression is not available. We use instead the \emph{KSG entropy estimate} based on $k$-nearest neighbor distances, as introduced in~\cite{kraskov2004estimating}. We assume that the CSGANs have captured the target posteriors well enough\textemdash based on the analysis in \S\ref{sec:oil_cgan_S} and \S\ref{sec:oil_cgan_f}\textemdash so that the CSGANs' uncertainty represents accurately the target posteriors' variance.

\begin{figure}[bt!]
    \centering
      \includegraphics[width=\textwidth]{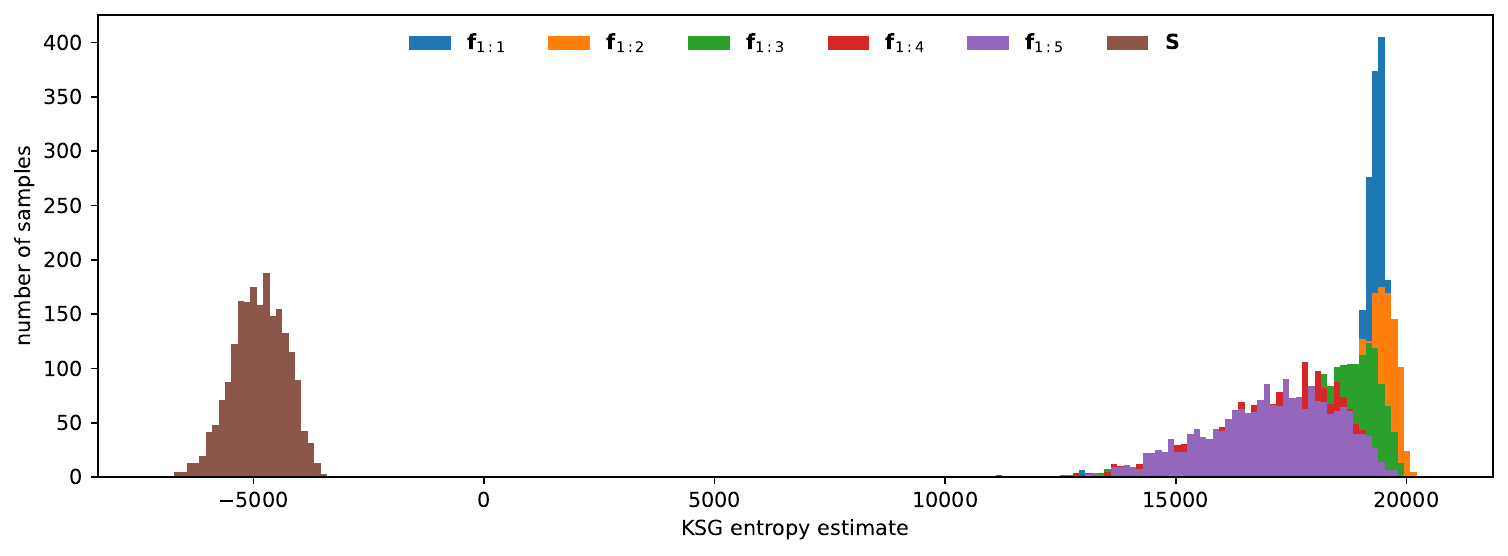}
    \\
      \includegraphics[width=0.19\textwidth]{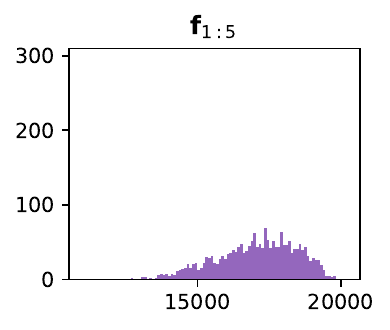}
      \hfill
      \includegraphics[width=0.19\textwidth]{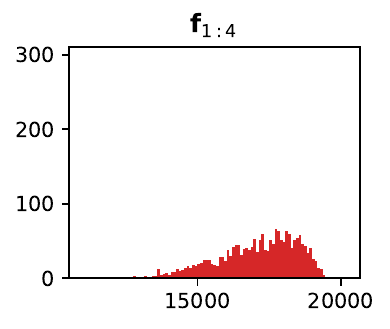}
      \hfill
      \includegraphics[width=0.19\textwidth]{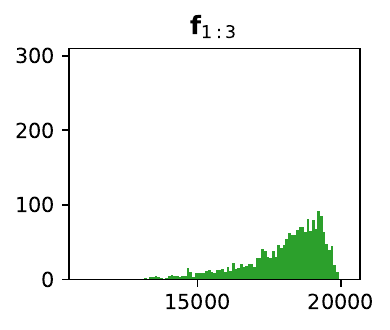}
      \hfill
      \includegraphics[width=0.19\textwidth]{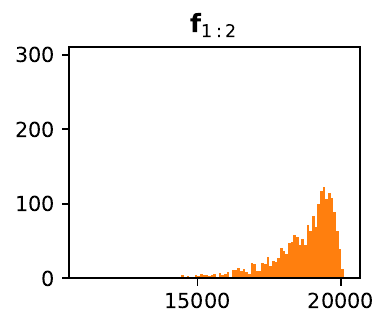}
      \hfill
      \includegraphics[width=0.19\textwidth]{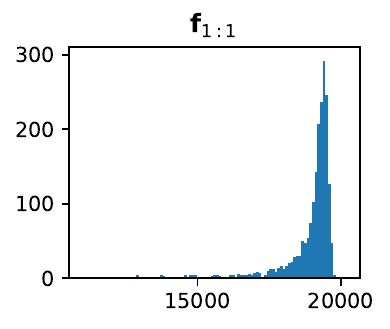}
  \caption{Entropy distribution of different latent CSGANs' predictions. Here each label refers to the input condition that the CSGAN takes, so as to identify each CSGAN.} 
  \label{fig:oil_etps}
\end{figure}

To assess the uncertainty of the six CSGANs that approximate $p(\mathbf{G}\mid \mathbf{S})$ and $p(\mathbf{G}\mid \mathbf{f}_{1:j}), j\in\{1,2,3,4,5\}$ respectively, and, more importantly, to put them on a unified scale for comparison, we first select 2000 test cases, then for each of them, we feed its input condition $(\mathbf{S}, \mathbf{f}_{1:j})$ to different CSGANs to let each CSGAN use different components of $(\mathbf{S}, \mathbf{f}_{1:j})$ to generate 1000 predictions $\hat{\mathbf{G}}$. Thereafter, for each CSGAN in each test case, we feed its 1000 predictions to the KSG estimator to evaluate the entropy of the distribution of $\hat{\mathbf{G}}$, so that after all evaluations, we eventually obtain a $6\times 2000$ grid of entropy estimates. Then we display the entropy distribution of the six CSGANs in Fig.~\ref{fig:oil_etps}.
The result reaffirms what we observed so far: the higher the intrinsic dimension of the input condition is, the lower the variance of the posterior. Figure~\ref{fig:oil_etps} shows that $p(\mathbf{G}\mid\mathbf{S})$ has much less entropy than $p(\mathbf{G}\mid\mathbf{f}_{1:j})$. Moreover, though being subjective, it appears that among $p(\mathbf{G}\mid\mathbf{f}_{1:j})$, the reduction in entropy starts to saturate as $j$ increases beyond 4, which aligns well with the saturation of $\mathbf{f}_{1:j}$'s intrinsic dimension as observed in \S\ref{sec:oil_lvdr}, Fig.~\ref{fig:oil_dr_f}.


\section{Conclusion}
\label{sec:conclusion}

We have thus far presented a holistic framework for Bayesian inverse problems in arbitrarily high dimensional scenarios and for an arbitrary complexity of the forward model that can be characterized by a highly nonlinear computer code. The proposed methodology leverages the novel Least Volume Autoencoders for identifying low dimensional manifolds on which the input and output data reside, and after encoding the observations, a conditional generative adversarial network is trained using a Sinkhorn divergence-based loss function for further improving its learning potential compared to its traditional GAN counterparts. On one hand, the proposed autoencoder structure used for dimensionality reduction allows working in extremely high dimensional settings as long as the data can be compressed to a low dimensional space. On the other hand, the GAN framework can tackle highly nonlinear forward models. For both tasks, all that is needed is to identify a rich enough architecture for the neural networks involved in order to guarantee an application-specific accurate inference procedure. 
The performance of the proposed approach is illustrated through two numerical examples, and the combination of computational efficiency along with the accuracy of the posterior solutions is unparalleled. We have provided a systematic study on how the quality of the posterior solution to the inverse problem improves as the intrinsic dimensionality of the observable quantity increases, thus providing additional information about the input, or, in other words, reducing the intrinsic dimensionality of the solution according to the preimage theorem. 

In the challenging setting of Example \ref{sec:oilextraction}, we have reached a limitation where the CSGAN cannot make consistently accurate prediction on different test cases when the observable quantity's intrinsic dimension is low (see \S\ref{sec:oil_cgan_f}). For complicated high dimensional inverse problems, there could be many elusive culprits that are hard to track down. However, we give three major hypotheses here to provide some insights on the limitations for arbitrarily complex experiments. 
\begin{itemize}
\item Our first hypothesis is inspired by our observation and discussion in \S\ref{sec:ko_cgan}. There, the low dimensional CSGAN learning the simpler bimodal distribution generates invalid predictions connecting the two modes (Fig.~\ref{fig:ko_cgan}), due to the \emph{intrinsic flaw} of pushforward models (see \S\ref{sec:ko_cgan}). The same issue can be possibly observed in the high dimensional cases of $p(\mathbf{G}\mid\mathbf{f}_{1:j})$ in \S\ref{sec:oil_cgan_f}. The preimage theorem states that under certain assumptions, the preimage of a given observed quantity is a submanifold of a certain dimension, which is usually conceived as a high dimensional curved surface in the inferred parameter space. Yet this surface can have complicated topology that the pushforward models cannot readily handle. For instance, it might constitute multiple disconnected components and thus can be intuitively similar to a union of several disjoint surfaces; or it is an enclosed surface like a sphere. In these cases, the pushforward model's intrinsic flaw could be triggered and would lead to inferior learning results. In contrast, conditioning the log-permeability on the oil saturation field in Example \ref{sec:oilextraction} has a much simpler, probably unimodal distribution, such that the preimage of $\mathbf{S}$ consists of only one point, so it does not suffer from this flaw (see \S\ref{sec:oil_cgan_S}). 
\item Our second hypothesis is that the mapping from log-permeability to the oil production curves might have large local Lipschitz constant in many regions, so that even if a CSGAN makes predictions very close to the target preimage of a given observed curve, the resulting posterior prediction may still be far away from the original observation due to this high sensitivity. If the local constant varies dramatically across different regions, it can lead to the large variation in prediction error. 

\item Our third hypothesis is data deficiency, which commonly plagues most data-driven learning tasks across all research fields. Although to date it is still tricky to determine the \emph{sample complexity}\textemdash\ie, how many data samples we need exactly for a given data-driven learning tasks to achieve satisfying accuracy, it is at least both theoretically~\cite{narayanan2010sample, narayanan2009sample} and empirically~\cite{pope2021intrinsic} discovered that it depends on the \emph{intrinsic dimension}, \emph{intrinsic volume} and \emph{curvature} of the dataset, whereas independent of the extrinsic dimension\textemdash\ie, the dimension of the dataset's ambient space. Therefore, learning $p(\mathbf{G}\mid\mathbf{f}_{1:j})$ could require more data than learning $p(\mathbf{G}\mid\mathbf{S})$, since the preimages of $\mathbf{f}_{1:j}$ have much larger intrinsic dimensions and probably much larger curvatures than $\mathbf{S}$'s. 

In addition, this implies for high dimensional problems, we can only \emph{heuristically} determine how many samples we need for each learning task, as the number of variables in the dataset\textemdash\ie, the extrinsic dimension\textemdash provides no useful information about sample complexity. However, when the data is abundant, sample complexity may not matter that much. For instance, in the future we may apply the methods in the paper to real-world applications where unlimited data can be collected easily from sensors rather than costly simulations.
\end{itemize} 
Nevertheless, at this moment, no certain conclusion can be drawn for this complicated high dimensional problem, and we leave further investigations to future research.

At last, the discerned correlation, investigated in \S\ref{sec:corr_dim}, supports our hypothesis in \S\ref{sec:outline_exp} based on the preimage theorem, and may inspire many novel applications in the future. For instance, we may use this information about dimensionality retrieved by LVAEs to anticipate which combination of observable variables provides the largest amount of information for inferring the unknown system parameters of interest, how certain the inference might be, whether the distribution of an array of sensors on a structure is optimal and so forth. However, it should be noted here that the latent dimension that an LVAE retrieves is an \emph{upper-bounding global estimate} of the dataset's true intrinsic dimension~\cite{qiuyi2024compressing}. In many situations, the gap between them may not be closed. For instance, when the given $n$-D data manifold cannot be embedded in $\mathbb{R}^{n}$ (\eg, it is a sphere, torus, Klein bottle, \etc.), or when the dataset is a union of several manifolds intersecting each other, such that some part of it is not locally Euclidean, or when the data manifold forms a nontrivial knot. We should always treat the LVAEs' results with these caveats in mind. 

\section*{Acknowledgements}

The information, work, or data presented herein was funded in part by the Advanced Research Projects Agency-Energy (ARPA-E), U.S. Department of Energy, under Award Number DE-AR0001295. The views and opinions of authors expressed herein do not necessarily state or reflect those of the United States Government of any agency thereof.

\appendix
\newpage

\section{Model Training Details of \S\ref{sec:ko_ode}}
Table~\ref{tab:ko_lvae} shows the architecture of the LVAE for $y_1$. Here all convolutional (Conv) and deconvolutional layers (Deconv) are 1D, and the prefix ``SN'' stands for ``spectral-normalized''. The subscript of each Conv, Deconv and Linear layer indicates the number of output channels/features. The Scale layer simply increases the network's Lipschtiz constant by scaling up each input neuron by the factor specified in its subscript, in order to accelerate the LVAE's convergence. All the LeakyReLU activations have $\alpha=0.2$.

\begin{table}[hbt!]
  \caption{Architecture of LVAE for $y_1$}
  \centering
  \label{tab:ko_lvae}
  \small
  \begin{tabular}{c|c|c}
    \toprule
    Input/Output & Encoder   &    Decoder \\
    \midrule
    \makecell[l]{$y_1\in \mathbb{R}^{256\times 1}$ \\ $z \in \mathbb{R}^{50}$}
    &
    \makecell[l]{$y_1 
    \to\text{Conv}_{32}\to\text{LeakyReLU}$ \\ 
    $\to\text{Conv}_{64}\to\text{LeakyReLU}$ \\ 
    $\to\text{Conv}_{64}\to\text{LeakyReLU}$ \\ 
    $\to\text{Conv}_{128}\to\text{LeakyReLU}$ \\ 
    $\to\text{Conv}_{256}\to\text{LeakyReLU}$ \\ 
    $\to\text{Reshape}_{8\times 256 \to 2048}$ \\ 
    $\to\text{Linear}_{256} \to \text{LeakyReLU}$\\
    $\to\text{Linear}_{128} \to \text{LeakyReLU}$\\
    $\to\text{Linear}_{50}\to z$} 
    &
    \makecell[l]{$z \to \text{Scale}_{256}$\\
    $\to\text{SN-Linear}_{128}\to\text{LeakyReLU}$ \\
    $\to\text{SN-Linear}_{256}\to\text{LeakyReLU}$ \\
    $\to\text{SN-Linear}_{2048}\to\text{LeakyReLU}$ \\
    $\to\text{Reshape}_{2048 \to 8\times 256}$ \\ 
    $\to\text{SN-Deconv}_{128}\to\text{LeakyReLU}$ \\ 
    $\to\text{SN-Deconv}_{64}\to\text{LeakyReLU}$ \\ 
    $\to\text{SN-Deconv}_{64}\to\text{LeakyReLU}$ \\ 
    $\to\text{SN-Deconv}_{32}\to\text{LeakyReLU}$ \\ 
    $\to\text{SN-Deconv}_{1}\to\text{Sigmoid} \to y_1$} \\ 
    \bottomrule
  \end{tabular}
\end{table}



\begin{table}[hbt!]
  \caption{Architecture of Latent CSGAN for $p(\bxi\mid y_1)$}
  \centering
  \label{tab:ko_csgan}
  \begin{tabular}{c}
    \toprule
    Generator   \\
    \midrule
    \makecell[l]{$z_{y_1}\times u \in \mathbb{R}^{7+2}$ \\ 
    $\to\text{Linear}_{512} \to \text{LeakyReLU}$\\
    $\to\text{Linear}_{512} \to \text{LeakyReLU}$\\
    $\to\text{Linear}_{64} \to \text{LeakyReLU}$\\
    $\to\text{Linear}_{2} \to \bxi\in\mathbb{R}^{2}$} \\
    \bottomrule
  \end{tabular}
\end{table}

\clearpage
\section{Model Training Details of \S\ref{sec:oilextraction}}

\begin{table}[h!]
  \caption{Architectures of LVAEs for $\mathbf{G}$, $\mathbf{S}$, $\mathbf{f}_{1:j}$}
  \centering
  \label{tab:oil_lvae}
  \small
  \begin{tabular}{c|c|c}
    \toprule
     Input/Output  &  Encoder   &    Decoder \\
    \midrule
    \makecell[l]{$\mathbf{G}\in \mathbb{R}^{128\times 64\times 1}$ \\
    $z\in \mathbb{R}^{100}$}
    & 
    \makecell[l]{$\mathbf{G}\to\text{Conv}_{64}\to\text{LeakyReLU}$ \\ 
    $\to\text{Conv}_{64}\to\text{LeakyReLU}$ \\ 
    $\to\text{Conv}_{128}\to\text{LeakyReLU}$ \\ 
    $\to\text{Conv}_{128}\to\text{LeakyReLU}$ \\ 
    $\to\text{Conv}_{256}\to\text{LeakyReLU}$ \\ 
    $\to\text{Reshape}_{4\times 2\times 256 \to 2048}$ \\ 
    $\to\text{Linear}_{1024} \to \text{LeakyReLU}$\\
    $\to\text{Linear}_{512} \to \text{LeakyReLU}$\\
    $\to\text{Linear}_{100}\to z$} 
    &
    \makecell[l]{$z \to \text{Scale}_{8192}$\\
    $\to\text{SN-Linear}_{512}\to\text{LeakyReLU}$ \\
    $\to\text{SN-Linear}_{1024}\to\text{LeakyReLU}$ \\
    $\to\text{SN-Linear}_{2048}\to\text{LeakyReLU}$ \\
    $\to\text{Reshape}_{2048 \to 4\times 2 \times 256}$ \\ 
    $\to\text{SN-Deconv}_{128}\to\text{LeakyReLU}$ \\ 
    $\to\text{SN-Deconv}_{128}\to\text{LeakyReLU}$ \\ 
    $\to\text{SN-Deconv}_{64}\to\text{LeakyReLU}$ \\ 
    $\to\text{SN-Deconv}_{64}\to\text{LeakyReLU}$ \\ 
    $\to\text{SN-Deconv}_{1}\to\text{Sigmoid} \to \mathbf{G}$ 
    } \\ 
    \midrule
    \makecell[l]{$\mathbf{S}\in \mathbb{R}^{128\times 64\times 1}$ \\
    $z\in \mathbb{R}^{100}$}
    & 
    \makecell[l]{$\mathbf{S}
    \to\text{Conv}_{64}\to\text{LeakyReLU}$ \\ 
    $\to\text{Conv}_{128}\to\text{LeakyReLU}$ \\ 
    $\to\text{Conv}_{256}\to\text{LeakyReLU}$ \\ 
    $\to\text{Conv}_{512}\to\text{LeakyReLU}$ \\ 
    $\to\text{Conv}_{1024}\to\text{LeakyReLU}$ \\ 
    $\to\text{Reshape}_{4\times 2\times 1024 \to 8192}$ \\ 
    $\to\text{Linear}_{1024} \to \text{LeakyReLU}$\\
    $\to\text{Linear}_{100}\to z$} 
    &
    \makecell[l]{$z \to \text{Scale}_{8192}$\\
    $\to\text{SN-Linear}_{1024}\to\text{LeakyReLU}$ \\
    $\to\text{SN-Linear}_{8192}\to\text{LeakyReLU}$ \\
    $\to\text{Reshape}_{8192 \to 4\times 2 \times 1024}$ \\ 
    $\to\text{SN-Deconv}_{512}\to\text{LeakyReLU}$ \\ 
    $\to\text{SN-Deconv}_{256}\to\text{LeakyReLU}$ \\ 
    $\to\text{SN-Deconv}_{128}\to\text{LeakyReLU}$ \\ 
    $\to\text{SN-Deconv}_{64}\to\text{LeakyReLU}$ \\ 
    $\to\text{SN-Deconv}_{1}\to\text{Sigmoid} \to \mathbf{S}$ 
    } \\ 
    \midrule
    \makecell[l]{$\mathbf{f}_{1:j}\in \mathbb{R}^{128\times j}$ \\
    $z\in \mathbb{R}^{50}$}
    & 
    \makecell[l]{$\mathbf{f}_{1:j} 
    \to\text{Conv}_{64j}\to\text{LeakyReLU}$ \\ 
    $\to\text{Conv}_{64j}\to\text{LeakyReLU}$ \\ 
    $\to\text{Conv}_{128j}\to\text{LeakyReLU}$ \\ 
    $\to\text{Conv}_{128j}\to\text{LeakyReLU}$ \\ 
    $\to\text{Reshape}_{8\times 128j \to 1024j}$ \\ 
    $\to\text{Linear}_{1024} \to \text{LeakyReLU}$\\
    $\to\text{Linear}_{512} \to \text{LeakyReLU}$\\
    $\to\text{Linear}_{512} \to \text{LeakyReLU}$\\
    $\to\text{Linear}_{50}\to z$} 
    &
    \makecell[l]{$z \to \text{Scale}_{128j}$\\
    $\to\text{SN-Linear}_{512}\to\text{LeakyReLU}$ \\
    $\to\text{SN-Linear}_{512}\to\text{LeakyReLU}$ \\
    $\to\text{SN-Linear}_{1024}\to\text{LeakyReLU}$ \\
    $\to\text{SN-Linear}_{1024j}\to\text{LeakyReLU}$ \\
    $\to\text{Reshape}_{1024j \to 8\times 128j}$ \\ 
    $\to\text{SN-Deconv}_{128j}\to\text{LeakyReLU}$ \\ 
    $\to\text{SN-Deconv}_{64j}\to\text{LeakyReLU}$ \\ 
    $\to\text{SN-Deconv}_{64j}\to\text{LeakyReLU}$ \\ 
    $\to\text{SN-Deconv}_{j}\to\text{Sigmoid}\to \mathbf{f}_{1:j}$
    } \\ 
    \bottomrule
  \end{tabular}
\end{table}

\begin{table}[h!]
  \caption{Hyperparameters of LVAEs for $\mathbf{G}$, $\mathbf{S}$, $\mathbf{f}_{1:j}$}
  \centering
  \label{tab:oil_lvae_hyp}
  \begin{tabular}{c|ccccc}
    \toprule
    & Batch Size & $\lambda$ & $\eta$ & Learning Rate (Adam) & Epochs\\
    \midrule
    $\mathbf{G}$ & 50    & $1$ & 1/8192 & 0.0001 & 6000 \\
    \midrule
    $\mathbf{S}$ & 50    & $5$ & 1/8192 & 0.0001 & 6000 \\
    \midrule
    $\mathbf{f}_{1}$ & 50    & $0.016$ & 1/128 & 0.0001 & 1000 \\
    \midrule
    $\mathbf{f}_{1:2}$ & 50    & $0.008$ & 1/128 & 0.0001 & 1000 \\
    \midrule
    $\mathbf{f}_{1:3}$ & 50    & $0.004$ & 1/128 & 0.0001 & 1000 \\
    \midrule
    $\mathbf{f}_{1:4}$ & 50    & $0.003$ & 1/128 & 0.0001 & 1000 \\
    \midrule
    $\mathbf{f}_{1:5}$ & 50    & $0.0025$ & 1/128 & 0.0001 & 1000 \\
    \bottomrule
  \end{tabular}
\end{table}


\begin{table}[h!]
  \caption{Architecture of Latent CSGANs for $p(\mathbf{G}\mid \mathbf{S})$ and $p(\mathbf{G}\mid \mathbf{f}_{1:j})$}
  \centering
  \label{tab:oil_csgan}
  \begin{tabular}{c|c}
    \toprule
    $p(\mathbf{G}\mid \mathbf{S})$ & $p(\mathbf{G}\mid \mathbf{f}_{1:j})$    \\
    \midrule
    \makecell[l]{$z_{\mathbf{S}}\times u \in \mathbb{R}^{40+40}$ \\ 
    $\to\text{Linear}_{512} \to \text{LeakyReLU}$\\
    $\to\text{Linear}_{512} \to \text{LeakyReLU}$\\
    $\to\text{Linear}_{256} \to \text{LeakyReLU}$\\
    $\to\text{Linear}_{40} \to z_\mathbf{G}\in\mathbb{R}^{40}$}
    &
    \makecell[l]{$z_{\mathbf{f}_{1:j}} \times u \in \mathbb{R}^{\dim \mathbf{f}_{1:j} +40}$ \\ 
    $\to\text{Linear}_{512} \to \text{LeakyReLU}$\\
    $\to\text{Linear}_{512} \to \text{LeakyReLU}$\\
    $\to\text{Linear}_{256} \to \text{LeakyReLU}$\\
    $\to\text{Linear}_{40} \to z_\mathbf{G}\in\mathbb{R}^{40}$}
    \\
    \bottomrule
  \end{tabular}
  \begin{tabular}{c}
        \footnotesize
       * $\dim \mathbf{f}_{1:j} = \{4, 8, 12, 16, 18\}$ for $j= \{1,2,3,4,5\}$ (see Fig.~\ref{fig:oil_lv_f})
  \end{tabular}
\end{table}

\begin{table}[hbt!]
  \caption{Hyperparameters of Latent CSGANs for $p(\mathbf{G}\mid \mathbf{S})$ and $p(\mathbf{G}\mid \mathbf{f}_{1:j})$}
  \centering
  \label{tab:oil_csgan_hyp}
  \begin{tabular}{c|cccc}
    \toprule
    & Batch Size & $\rho$ of $\mathrm{S}_\rho$ & Learning Rate (Adam) & Epochs\\
    \midrule
    $\mathbf{S}$ & 128    & $0.3$ & 0.0001 & 2000 \\
    \midrule
    $\mathbf{f}_{1:j}$ & 128    & $0.3$ & 0.0001 & 3000 \\
    \bottomrule
  \end{tabular}
\end{table}

\section{Tips for Hyperparameter Tuning}
The training of the LVAEs involves the hyperparameters $\lambda$, $\eta$ and the decoder's Lipschitz constant $K$. A comprehensive guideline for tuning them can be found in Appendix C of~\cite{qiuyi2024compressing}.

The training of the Sinkhorn GANs requires setting the hyperparameter $\rho$ for the Sinkhorn divergence $\mathrm{S}_\rho$. A detailed study of its effect can be found in~\cite{genevay2018learning, feydy2019interpolating, genevay2019sample}. In brief, a smaller $\rho$ offers $\mathrm{S}_\rho$ better resolution to differentiate between different data samples, thus leads to better alignment of the generator $g$'s image with the target dataset (see examples in~\cite{feydy2019geometric}), but it incurs worse sample complexity and lower Sinkhorn algorithm convergence speed. So $\rho$ makes trade-off between quality and efficiency. 
The $\rho$ in Table~\ref{tab:oil_csgan_hyp} were not tuned meticulously. We basically applied $\rho=0.3$ to the $p(\mathbf{G}\mid \mathbf{S})$ case at first and obtained ideal results, then we straightforwardly applied it to all the experiments of $p(\mathbf{G}\mid \mathbf{f}_{1:j})$ without further change. If the precision of Sinkhorn GANs in approximating the target distribution $\mathbb{P}_r$ is the primary concern in certain applications, $\rho$ may be fine-tuned against a certain statistical divergence $D(\mathbb{P}_g, \mathbb{P}_r)$ independent of $\rho$, \eg, the MMD~\cite{smola2007hilbert}.




\end{document}